\documentclass{article}

\usepackage{xcolor}
\usepackage{amsmath}
\usepackage{amssymb}
\usepackage{amsfonts}
\usepackage{subcaption}
\usepackage{graphicx}
\usepackage{booktabs}
\usepackage{tabularx}
\usepackage{multirow}
\usepackage{xspace}
\usepackage{enumitem}
\usepackage{svg}
\usepackage{placeins}
\usepackage{tikz}
\usetikzlibrary{shapes.geometric, arrows.meta, positioning, calc}

\usepackage[final]{corl_2026} % Uncomment 

\newcommand{\method}{STARS\xspace}
\newcommand{\methodfull}{\textbf{S}patio\textbf{T}emporal \textbf{A}utoencoded \textbf{R}epresentations for \textbf{S}ocial-interaction\xspace}

\title{STARS: From Spatiotemporal Dynamics to Social Representations in Human-Robot Interaction}

\author{
\textbf{Nathan Tsoi\textsuperscript{*},  
Michael J.~Munje\textsuperscript{*},  
Tejas Oberoi, 
Rishab Maheshwari,}\\
\textbf{Pengen Zheng, 
Tanush Chauhan, 
Peter Stone,
Joydeep Biswas}\\
\textsuperscript{}Department of Computer Science, The University of Texas at Austin \\
\textsuperscript{*} Equal Contribution 
}

\begin{document}
\maketitle
% \begingroup
% \renewcommand{\thefootnote}{\ensuremath{}}
% \footnotetext{Correspondence to: \texttt{nathan.tsoi@utexas.edu}.}
% \endgroup

\begin{abstract}
Fielding socially competent robots requires joint reasoning over spatial information and the social information it can convey, such as group motion and personal space. While current models excel at either spatial reasoning (e.g., modeling physical dynamics) or social reasoning (e.g., parsing text-based interactions), few natively unify both. Moreover, data scarcity in Human-Robot Interaction (HRI) makes training large models from scratch challenging. To bridge this gap, we introduce \textbf{STARS}: \textbf{S}patio\textbf{T}emporal \textbf{A}utoencoded \textbf{R}epresentations for \textbf{S}ocial-interaction, a method for learning \emph{event-level relational representations} that models short windows of multi-agent interactions as spatiotemporal graphs. A self-supervised message-passing graph neural network autoencoder then compresses these physical dynamics into a compact latent space for each node. By explicitly modeling agents and their interactions as graph structure, STARS injects a strong relational inductive bias to naturally capture the underlying social situation. To assess the utility of these learned representations, we evaluate STARS' representations on the SEAN Together dataset, which provides VR-based navigation data annotated with subjective human perceptions, and the SocialNav-SUB benchmark for social scene understanding. When trained on STARS representations, linear probes achieve highly competitive performance on perception prediction and pedestrian action classification tasks. Specifically, STARS exhibits strong data efficiency, particularly on pedestrian action classification, where \method-He consistently outperforms unstructured baselines across labeled-data fractions, including in low-data settings. Furthermore, qualitative analysis of the learned latent space reveals that STARS representations can capture high-level social semantics, demonstrating cohesive clustering of distinct human action intents without optimizing the encoder with a supervised action classification objective. An overview of this paper along with the code can be found at \url{https://larg.github.io/stars}.
\end{abstract}

\keywords{Human-Robot Interaction, Social Navigation, Representation Learning, Graph Neural Networks} 

\section{Introduction}
Deploying socially competent robots in human spaces requires autonomous robots to reason jointly using spatial knowledge to plan collision-free paths and social knowledge to perceive context, predict behaviors, and adhere to social norms. A robot navigating through a crowded hallway, for example, must plan collision-free motion in the physical environment while also anticipating pedestrian behavior and reacting in a socially compliant manner~\cite{mavrogiannis2023core, francis2025principles}.
Joint reasoning over spatial and social information is central to Human-Robot Interaction (HRI), yet existing approaches tend to address only one or the other.
Foundation spatial models parse LiDAR and visual observations into geometric representations, but lack the capacity to understand social dynamics~\cite{zhou2025social, lisondra2026embodied}.
At the same time, some autoregressive models conditioned on text (LLMs) or text and vision (VLMs) exhibit strong social perception~\cite{zhou2025social}, but struggle to process continuous spatial dynamics~\cite{ramakrishnan2024does, munje2025socialnavsub} and precise temporal information~\cite{gao2025vision}.
Bridging this gap by training large models from scratch is challenging due to the scarcity of grounded HRI data.
Collecting such data requires time and cost intensive human-subject studies, and yields datasets that are small, fragmented across sensor modalities, and highly prone to overfitting when applied to large, unstructured models.

Although most HRI datasets are small and fragmented, often collected in different environments, with different robots, sensors, tasks, and annotations, the premise of this paper is that their recurring social interaction structure~\cite{de2015makes} can be used to learn reusable representations for downstream social navigation tasks that require reasoning about agents, interactions, and shared context. This interaction structure is naturally relational, as agents move through shared spaces and influence one another over time. We hypothesize that graph-structured representations are well-suited to learning from small, fragmented HRI datasets because they can encode agents as nodes and interactions as edges, injecting strong \textit{relational inductive bias}~\cite{battaglia2018relational}. This inductive bias enables message-passing graph neural networks~\cite{gilmer2020message} to learn interaction patterns efficiently in limited-data regimes.

We introduce \textbf{\method}: \methodfull, a method for self-supervised learning of \emph{event-level relational representations}.
In contrast to the autoregressive prediction of world models, \method learns a representation over a short time horizon to capture relevant contextual information.
We define an \emph{interaction scenario} as a fixed-duration sequence of events where agents share a physical space.
We encode these interaction scenarios into graphs, where directed edges encode sequences of pairwise spatial features over time.

Though our interaction scenario graphs contain only spatiotemporal information, our target downstream tasks are inherently social.
By optimizing a self-supervised reconstruction objective over the fixed-duration interaction scenario, \method compresses physical dynamics into a compact latent space that captures the implicit \textit{social situation}~\cite{tsoi2022sean}.
We demonstrate our method using two architectural variants that show the learned representations allow a simple linear probe to extract social meaning from physical motion, unlocking data-efficient generalization without requiring task-specific representation learning.

Our contributions are threefold: 
\vspace{-0.5em}
\begin{enumerate}[nosep, leftmargin=*]
    \item We introduce \method, a self-supervised representation learning method for social navigation, which compresses datasets with different feature dimensions and graph schemas into a latent representation containing features useful for diverse downstream tasks with limited task-specific labeled data.
    \item We demonstrate that self-supervised pretraining of relational representations improves data efficiency. By comparing \method to unstructured baselines in low-data regimes, we show that pretrained graph representations support robust downstream generalization with a fraction of the labeled data and transfer across the two evaluated datasets.
    \item We establish that \method' self-supervised representations naturally encode social semantics. By analyzing the latent space topology, we illustrate the model's inherent ability to distinguish complex social behaviors, such as action intents, without a supervised action classification objective.
\end{enumerate}

\section{Related Work}
\label{sec:relatedwork}

\paragraph{Spatial and social reasoning in social navigation.}

Evaluating multi-agent interactions has historically relied on pedestrian trajectory datasets~\cite{pellegrini2009you, lerner2007crowds, robicquet2016learning} and egocentric robot observations~\cite{martin2019jrdb, karnan2022socially}.
To model these complex shared environments, Graph Neural Networks (GNNs) have emerged as the standard architecture, as they naturally mirror the topological structure of human crowds~\cite{mohamed2020social, kosaraju2019social}.
Recent heterogeneous and attention-based spatiotemporal GNNs excel at weighting relational dependencies between diverse entities to extract continuous kinematic interactions~\cite{li2025pedestrian, li2025unified}.
Current learning efforts tend to focus on trajectory forecasting, which aims to minimize spatial displacement error, and therefore learn to reflect future spatial configurations, rather than reusable social semantics.
Recent works have laid the groundwork to push beyond spatial forecasting.
\citet{zhang2023predicting} provide a dataset and baseline for predicting human perceptions of robot performance and \citet{munje2025socialnavsub} established a benchmark for VLM understanding of social interactions in navigation.
Still, existing trajectory-focused models struggle to generalize to \textit{social} tasks because they do not explicitly model the normative behaviors and interpersonal associations that define social interactions~\cite{thompson2025social, tsoi2022sean}.

\paragraph{Foundation models and the HRI data bottleneck.}
Foundation and world models demonstrate impressive predictive capabilities for embodied decision-making~\cite{bommasani2021opportunities, hafner2023mastering, bruce2024genie, parkerholder2024genie2}, including strong zero-shot transfer in Vision-Language-Action (VLA) controllers~\cite{o2024open, zitkovich2023rt, kim2024openvla}.
However, these models rely on massive, unified datasets.
In contrast, grounded HRI data is scarce, geographically constrained, and highly fragmented across varying sensor modalities, survey instruments, and tasks~\cite{duncan2024survey, ondras2022human, bu2024ssup, heinisch2024physiological,heo2024diverse, thompson2021conversational, yang2021dataset, shrestha2024natsgd}.
Applying foundation-scale learning to HRI is therefore hindered by data scarcity, leading unstructured models to overfit or fail to generalize across diverse social scenarios~\cite{lisondra2026embodied}.

\paragraph{Relational inductive bias.}
To learn efficiently in low-data regimes, recent approaches have leveraged structured representations.
Relational inductive bias~\cite{battaglia2018relational} constrains neural architectures to represent entities as nodes and interactions as edges, encouraging networks to deduce underlying physical dynamics without relying on hand-engineered features~\cite{ferreira2019learning}.
While this bias has driven advancements in physical object manipulation, its application to extracting social context from interaction remains underexplored~\cite{sanchez2020learning, malik2023relational}.
At the same time, generative self-supervised graph representation learning has shown that semantically meaningful embeddings can be extracted without human annotation by reconstructing masked features or topological structures~\cite{kipf2016variational, liu2025network, falqueto2025human}.

\section{Method}

In this section, we detail the \method framework.
We first formalize the problem of extracting representations from heterogeneous social navigation datasets, outline our graph construction process, and describe the unsupervised graph autoencoder used to learn the latent social semantics.

\label{sec:method}
\subsection{Problem Formulation}

Let $\mathcal{D} = \{\mathcal{D}_1, \mathcal{D}_2, \dots, \mathcal{D}_K\}$ denote a collection of social navigation datasets (Fig.~\ref{fig:ds_sns}A), where index $k$ identifies a dataset in the collection.
% For the purposes of this work, we consider an HRI dataset to be any dataset containing observations of humans and robots, or humans in shared physical spaces relevant to robot interaction.
Each dataset $\mathcal{D}_k$ contains a set of interaction scenarios $\{x^{(k)}_i\}$.
An interaction scenario $x$ is a fixed-duration observation of $T_k$ timesteps containing multiple agents sharing a physical space.
Rather than operating directly on arbitrary raw sensor inputs, \method converts each scenario into a common graph representation with agent node attributes and temporal pairwise edge features.
Let $\mathcal{A}$ denote the finite set of node types, $d_v^{(k)}$ the dimension of the node attributes supplied by dataset $k$, and $d_e$ the dimension of the pairwise features.
We represent each scenario with $N$ nodes as a tuple $G = (\boldsymbol{\tau}, \mathbf{X}^{V}, \mathbf{m}, \mathbf{X}^{E})$ in the graph space $\mathcal{G}_k$, defined as:
\begin{equation}
\label{eq:scenario_graph}
\mathcal{G}_k = \bigsqcup_{N \geq 2}\;
\underbrace{\mathcal{A}^{N}}_{\text{types } \boldsymbol{\tau}}
\;\times\;
\underbrace{\mathbb{R}^{N \times d_v^{(k)}}}_{\text{node attributes } \mathbf{X}^{V}}
\;\times\;
\underbrace{\{0,1\}^{N \times T_k}}_{\text{valid mask } \mathbf{m}}
\;\times\;
\underbrace{\mathbb{R}^{N \times N \times T_k \times d_e}}_{\text{edge features } \mathbf{X}^{E}}
\end{equation}
The disjoint union over $N$ allows the number of agents to vary across scenarios.
Each node $v_i \in V$ corresponds to an agent and has type $\tau_i \in \mathcal{A}$.
The node set is partitioned as $V = V_R \cup V_H$, where $V_R$ contains robot nodes and $V_H$ contains human nodes, and the type of each directed edge follows from its endpoints, $\rho_{ij} = (\tau_i, \tau_j)$.
The valid mask entry $m_{i,t}$ records whether node $i$ is observed at timestep $t$, and edge validity is given by $M^{E}_{ij,t} = m_{i,t}\,m_{j,t}\,\mathbb{1}[i \neq j]$, so the graph is complete over every ordered pair of nodes observed at the same timestep.
Each row $\mathbf{x}_i$ of $\mathbf{X}^{V}$ contains the node attributes available for agent $i$, such as encoded occupancy grids or trajectory goals, and is zero-initialized otherwise.
Each entry $\mathbf{e}_{ij} \in \mathbb{R}^{T_k \times d_e}$ of $\mathbf{X}^{E}$ is a temporal edge feature sequence containing the relative $SE(2)$ pose of node $j$ in the body frame of node $i$ and its temporal deltas, with $d_e = 8$ (Appendix~\ref{sec:edge_features}).
Datasets may differ in sensor modality, sampling frequency, agent types, and available annotations and in Eq.~\ref{eq:scenario_graph} that variation is confined to $d_v^{(k)}$ and $T_k$.
To handle these diverse inputs, all raw sensor modalities are projected into a common hidden dimension prior to message passing. 
The relational message-passing core and latent heads are shared across datasets, while dataset-dependent input projections, temporal edge encoders, and edge decoders accommodate differences in feature dimensions and temporal windows (Appendix~\ref{sec:sharing}).
 
The representation-learning problem maps an interaction scenario $x$, through its graph $G \in \mathcal{G}_k$, to compact per-node representations $\mathbf{Z}$ that support downstream social navigation tasks.
For a downstream task $\mathcal{T}_m$, a subset of scenarios is paired with labels $\{(x_i, y_i^{(m)})\}$,
where $y_i^{(m)} \in \mathcal{Y}_m$ represents the task-specific annotation, which in our evaluations include subjective human perceptions of robot behavior and discrete pedestrian action categories.
A representation is considered better if it yields stronger held-out performance on $\mathcal{T}_m$ under the same labeled-data budget and downstream adaptation protocol.
Thus, the goal is to obtain compact and reusable representations that support strong downstream performance when labeled examples for each task are scarce.

\begin{figure}[p!tb]
    \centering
    \includegraphics[width=\textwidth]{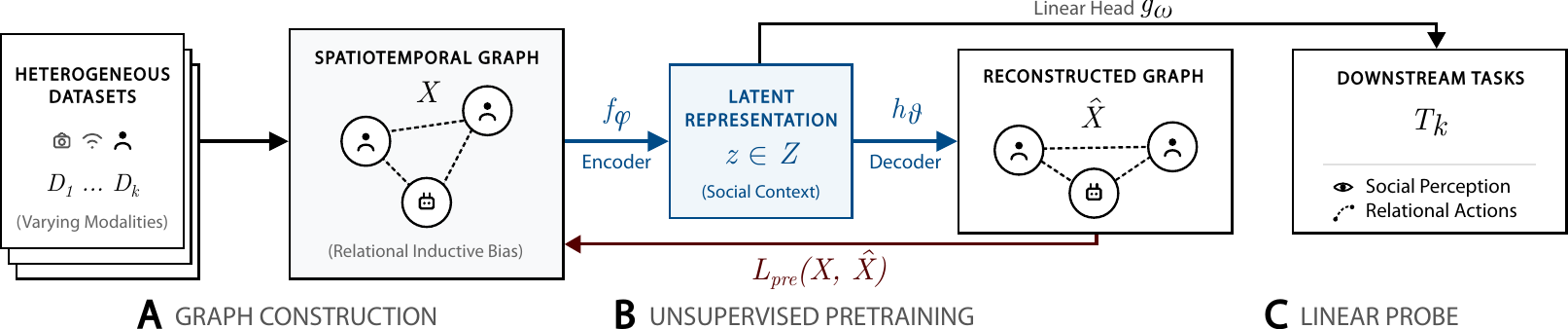}
    \caption{
    \textbf{\method} three stages:
    (\textbf{A}) construct spatiotemporal graphs from social navigation datasets,
    (\textbf{B}) utilize an unsupervised message-passing graph autoencoder to compress physical dynamics into latent representations, and
    (\textbf{C}) adapt frozen latents to downstream tasks using lightweight task-specific prediction heads.
    See Sec. \ref{sec:method} for details.
    }
    \label{fig:method}
\end{figure}

\subsection{\method Overview} 
Motivated by a need to learn from diverse social navigation datasets containing spatiotemporal interactions while preserving the relational structure needed for downstream social tasks, \method is 
designed to disentangle spatiotemporal dynamics via autoencoding to per-node latents $\mathbf{Z} = f_\phi(G)$ that capture reusable social context~\cite{tsoi2022sean}.

As shown in Figure~\ref{fig:method}, \method is a three-stage framework for learning representations from spatiotemporal data.
First, \method extracts features from social navigation datasets (examples shown in Figure~\ref{fig:ds_sns}) to construct unified spatiotemporal graphs.
While \method can apply generally to any spatiotemporal relational data, in this work we focus on the application area of social navigation.
Explicitly modeling human and robot agents as nodes and their physical interactions as edges imposes a strong relational inductive bias.
Second, a pretraining step compresses physical dynamics into a compact latent space via a self-supervised message-passing graph autoencoder.
Self-supervised reconstruction reduces overfitting to scarce labels, enabling higher-capacity encoders compared to a supervised learning baseline (Appendix~\ref{sec:supervised_heterograph_ablation}).
To evaluate the utility of these representations for social tasks, we subsequently apply a lightweight, task-specific linear prediction head to the frozen latents.
We instantiate \method using both homogeneous and heterogeneous graph formulations to study how explicit entity and relation types affect \method' learned representations.

\subsection{Spatiotemporal Relational Graph Construction}

Our construction makes two design choices.
First, our graphs are \textit{edge-feature dense}, capturing temporal relationships primarily through edge rather than node attributes.
Second, rather than using distance-based thresholds, we construct \textit{complete graphs} to capture interactions between all observed agents.
An example graph construction is shown in Appendix Fig.~\ref{fig:heter_architecture}, and full input projection specifications are given in Appendix~\ref{sec:stems}.
 
\textbf{The homogeneous graph} represents nodes as generic ``agents'' and all edges as ``interactions.''
Robot and human nodes retain distinct input attributes, and entity type enters only through a two-class node embedding; the graph does not include edge-type embeddings.
Every dataset is resampled to a common temporal window, so one input projection and one edge encoder serve both datasets, and minibatches may mix sources.
 
\textbf{The heterogeneous graph}
explicitly distinguishes nodes by entity type.
Edges are also typed based on the connected node types.
Robot nodes may include robot-specific attributes, such as encoded occupancy grids or trajectory goals when available, while human nodes are associated with a learnable semantic embedding representing the entity type, concatenated with a mean-pooled summary of their incident edge features.
The edge set captures typed relations such as robot-human, human-robot, and human-human interactions, with relation types provided separately through edge-type embeddings.
% Each dataset retains its native temporal window, so there is one edge encoder per dataset and one input projection per dataset and node type, and batches are restricted to a single source.

\subsection{Unsupervised Graph Autoencoder}

The core of \method is a graph encoder implemented as a Message-Passing Graph Neural Network (MPGNN) that maps the spatiotemporal graph into node-level latent distributions.

\textbf{Message Passing:}
The graph encoder performs message passing over the complete interaction graph.
In \method' \textit{heterogeneous} graph formulation, message passing uses a HEAT-style update~\cite{mo2021heat}.
Let $\mathbf{h}_j^{(\ell)}$ and $\mathbf{h}_i^{(\ell)}$ denote the source and destination node embeddings at layer $\ell$, and $\mathbf{t}_{ij}$ denote a learned embedding for the specific edge type (e.g., human-robot vs. human-human).
Let $\tilde{\mathbf{e}}_{ij}^{(0)} = \text{Enc}_{\text{edge}}(\mathbf{e}_{ij})$ denote the initial encoded temporal edge representation.
We construct a joint interaction context:
\begin{equation}
\mathbf{u}_{ij}^{(\ell)} =
[\mathbf{h}_j^{(\ell)} \| \mathbf{h}_i^{(\ell)} \| \tilde{\mathbf{e}}_{ij}^{(\ell)} \| \mathbf{t}_{ij}]
\end{equation}
The representation $\mathbf{u}_{ij}^{(\ell)}$ includes the full context of the interaction.
During message passing, the network must determine both the content of the information being transmitted and the relative importance of that information.
Therefore, we use $\mathbf{u}_{ij}^{(\ell)}$ to independently parameterize both the raw attention score $s_{ij}^{(\ell)}$ and the message content $\mathbf{m}_{ij}^{(\ell)}$ via two learnable functions:
\begin{equation}
s_{ij}^{(\ell)} = f_{\mathrm{attn}}(\mathbf{u}_{ij}^{(\ell)}), \qquad \mathbf{m}_{ij}^{(\ell)} = f_{\mathrm{msg}}(\mathbf{u}_{ij}^{(\ell)})
\end{equation}
Attention weights are normalized via a destination-wise softmax $\alpha_{ij}^{(\ell)}$, and the aggregated message is used to update the node states via a residual LayerNorm step. 
The heterogeneous formulation also updates $\tilde{\mathbf{e}}_{ij}^{(\ell)}$ between message-passing layers using the corresponding edge message; full update equations are provided in Appendix~\ref{sec:message-passing}.
Typed-edge attention enables the model to learn which distant interactions matter without needing separate message-passing modules per relation.
The \textit{homogeneous} graph formulation omits edge-type embeddings $\mathbf{t}_{ij}$ from $\mathbf{u}_{ij}^{(\ell)}$, using only the source node embedding, destination node embedding, and encoded temporal edge features $\tilde{\mathbf{e}}_{ij}$ in the update.

\textbf{Latent Encoding:} After $L$ message-passing layers, shared linear projection heads parameterize a diagonal Gaussian latent distribution for each node:
\begin{equation}
\label{eq:latent}
q_\phi(\mathbf{z}_i \mid G) = \mathcal{N}\!\left(\boldsymbol{\mu}_i,\; \mathrm{diag}(\boldsymbol{\sigma}_i^2)\right),
\qquad
\boldsymbol{\mu}_i = f_\mu\bigl(\mathbf{h}_i^{(L)}\bigr),
\quad
\log \boldsymbol{\sigma}_i^2 = f_\sigma\bigl(\mathbf{h}_i^{(L)}\bigr)
\end{equation}
Latent samples are drawn via the reparameterization trick, $\mathbf{z}_i = \boldsymbol{\mu}_i + \boldsymbol{\sigma}_i \odot \boldsymbol{\epsilon}$.
Let $\mathbf{Z} = [\mathbf{z}_1,\dots,\mathbf{z}_N]^\top \in \mathbb{R}^{N \times d_z}$ denote the stacked node latents, where $d_z$ is the latent dimension, so that $q_\phi(\mathbf{Z} \mid G) = \prod_i q_\phi(\mathbf{z}_i \mid G)$.
This per-node latent formulation preserves the asymmetry of pairwise interactions, as edge reconstruction conditions on both endpoint latent nodes rather than a single pooled graph representation.

\textbf{Decoding \& Objective:} Unlike autoregressive models, we employ an unsupervised autoencoding objective over the entire fixed-duration interaction scenario.
Specifically, for temporal edge reconstruction, the decoder employs an MLP that takes the concatenated source and destination latents together with the edge-type embedding, $[\mathbf{z}_i \| \mathbf{z}_j \| \mathbf{t}_{ij}]$, as input.

% This MLP projects the joint latent representation into a hidden dimension and outputs a single flat vector of $T_k d_e$ values for each edge.
% This sequence-level output is then reshaped into the explicit temporal sequence format $T_k \times d_e$ to yield the reconstructed edge features $\mathbf{\hat{e}}_{ij}$; because the output dimension is set by the window, the edge decoder is instantiated per dataset.
 
For node-level reconstruction, we restrict the objective to nodes containing active input attributes. 
Reconstructing static, zero-initialized features, such as placeholders for entities without observed attributes, would introduce a strong bias toward trivial constant targets, diluting the gradient signal.
Let $V_{\text{feat}} \subseteq V$ be the subset of nodes containing active attributes, determined by the dataset and node type rather than by testing whether $\mathbf{x}_i$ is zero (Appendix~\ref{sec:stems}).
In practice, $V_{\text{feat}}$ includes only nodes with observed, non-placeholder attributes, which in our datasets is the SEAN-T robot node alone.
For nodes without active input attributes, we omit node reconstruction. Table~\ref{tab:features_training} in Appendix~\ref{sec:stems}
summarizes the node features and reconstruction targets for each
architecture and dataset.
Crucially, the absence of human node reconstruction does not exclude human nodes from the learned representations or lead to a training mismatch, because the temporal edge decoder must reconstruct $\mathbf{\hat{e}}_{ij}$ from the concatenated endpoint latents alone.
The model is forced to encode rich spatiotemporal social semantics and joint context into the human node latents to successfully reconstruct these edges.
 
The model is trained by minimizing the following VAE objective:
\begin{equation}
\label{eq:elbo}
\mathcal{L}_{\text{pre}}
= \lambda_{v} \!\!\sum_{v_i \in V_{\text{feat}}}\!\! \text{MSE}(\mathbf{x}_i, \mathbf{\hat{x}}_i)
\;+\; \sum_{\rho} \lambda_{\rho} \!\!\sum_{\rho_{ij} = \rho}\!\! \text{MSE}(\mathbf{e}_{ij}, \mathbf{\hat{e}}_{ij})
\;+\; \beta \sum_{v_i \in V} D_{\text{KL}} \left( q_\phi(\mathbf{z}_i \mid G) \parallel \mathcal{N}(0, I) \right)
\end{equation}
where $\lambda_{v}$ and $\lambda_{\rho}$ denote weighting factors for node and edge reconstruction, the latter indexed by relation type, allowing the network to learn across imbalanced datasets, for example, where human-human interactions are much more common than human-robot interactions.

\subsection{Task-Agnostic Linear Probing}

To evaluate the quality of our learned latent representations and adapt them to downstream tasks with limited labeled data, we apply linear probes~\cite{alain2016understanding}.

Because \method learns compressed node-level latent representations $\mathbf{z}_i$, it provides flexibility in selecting and aggregating these node latents to represent different structural components relevant to a given downstream task.
After freezing the pretrained encoder weights $\phi$, \method selects and aggregates the necessary latents based on a given downstream task, and because $N$ varies across scenarios this aggregation returns a fixed-dimensional vector that is invariant to permutations of nodes of the same type:
\begin{equation}
\label{eq:readouts}
\mathbf{z}^{\text{SEAN}}_{\text{probe}} = \bigl[\, \mathbf{z}_{r} \,\big\|\, \overline{\mathbf{z}}_{h} \,\bigr],
\qquad
\overline{\mathbf{z}}_{h} = \tfrac{1}{|V_H|}\!\!\sum_{v_i \in V_H}\!\! \mathbf{z}_i,
\qquad
\mathbf{z}^{\text{SNS}}_{\text{probe}}(i) = \bigl[\, \mathbf{z}_{r} \,\big\|\, \mathbf{z}_i \,\bigr],
\quad v_i \in V_H
\end{equation}
the first for the graph-level SEAN-T perception tasks and the second for the node-level SNS action task, both of dimension $2 d_z$.
A linear probe with weights $W$ is then optimized using supervised pairs $(G, y)$, where $y$ is the task label, by minimizing $\mathcal{L}_{\text{task}} = \text{LossFn}(y, W \mathbf{z}_{\text{probe}} + b)$.
This lightweight adaptation step keeps the representation fixed, so downstream performance reflects the information encoded in $\mathbf{z}_{\text{probe}}$ rather than task-specific representation learning.

% Source is in figures/datasets.kra (a https://krita.org/en/ app file)
\begin{figure*}[t]
  \centering
  \includegraphics[width=\linewidth]{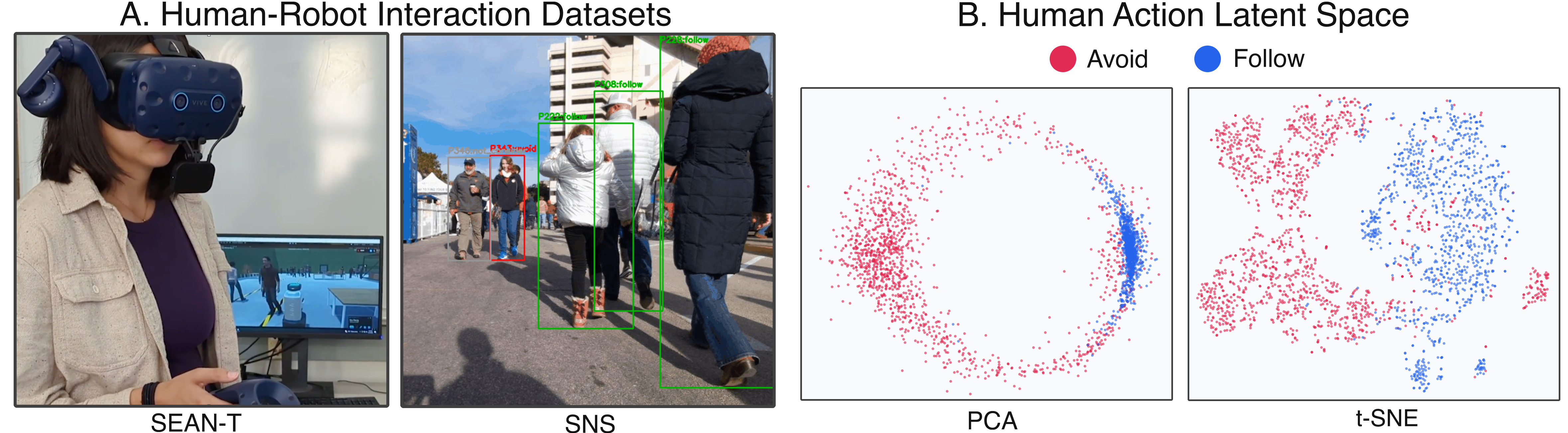}
  \caption{\textbf{(A) Dataset Overview.} 
  We utilize two human-robot interaction datasets: \textbf{SEAN-T}~\cite{zhang2023predicting} for evaluating subjective human ratings of robot navigation and \textbf{SNS}~\cite{munje2025socialnavsub} for downstream relational action classification (e.g., \textit{avoid}, \textit{follow}). 
  \textbf{(B) Latent Space Analysis (RQ3).}
  2D Projections of the SNS Human Node Action Latent Space for the \textit{avoid} and \textit{follow} actions.
  We visualize the 128-dimensional representations of nodes extracted from the \method-trained 
  homogeneous GNN-VAE encoder using linear \textbf{PCA} and non-linear \textbf{t-SNE}.
  Each point represents a node embedding, color-coded by action label.
  See Sec.~\ref{sec:latent_space_semantics} for details.}
  \label{fig:ds_sns}
\end{figure*}

\section{Experiments}
\label{sec:experiments}

We evaluate the effectiveness of \method to learn social semantics from physical dynamics.
Specifically, we demonstrate that \method achieves data-efficient downstream performance, and through analysis of the latent space topology, we show that the model can distinguish social behaviors, such as action intents, without optimizing the encoder with a supervised classification objective during pretraining.
Our research questions are:

\begin{enumerate}[nosep, label={\textbf{RQ\arabic*:}}, leftmargin=*]
    \item \textbf{Downstream Performance.} How well does \method work on downstream social navigation tasks compared to unstructured baselines when adapting pretrained representations?
    \item \textbf{Data Efficiency.} For a given downstream task, what is the impact of labeled data availability, from 1\% of the labeled training set (60 samples) to 100\%, when using pretrained \method representations?
    \item \textbf{Latent Space Semantics.} Do \method' self-supervised representations naturally encode social semantics (e.g., action intents) without a supervised action classification objective?
\end{enumerate}

\paragraph{Datasets and Tasks}
We evaluate \method across two distinct datasets to assess unsupervised pretraining and downstream task adaptation.
Our pretraining pool consists of multi-agent spatiotemporal data from \textbf{SEAN Together (SEAN-T)}~\cite{zhang2023predicting}, which provides 8-second immersive virtual-reality observations of robot navigation, and \textbf{SocialNav-SUB (SNS)}~\cite{munje2025socialnavsub}, a structured social reasoning benchmark containing pedestrian-robot interactions.
Examples of the interaction scenarios are shown in Figure~\ref{fig:ds_sns} and further details are provided in the Appendix~\ref{sec:datasets}.
To evaluate whether our autoencoder successfully captures reusable social semantics, we freeze the pretrained representations and train lightweight linear probes on two distinct downstream tasks: 
1) \textbf{Perception Prediction (SEAN-T):} Predicting binary subjective human ratings of a robot's behavior across three distinct metrics: perceived competence, clarity of intention, and surprise~\cite{zhang2023predicting}. This task evaluates the model's ability to extract subjective social impressions directly from physical motion.
2) \textbf{Action Classification (SNS):} Predicting discrete, categorical relational actions taken by pedestrians in dense crowd scenes with up to 30 people (e.g., \emph{avoiding}, \emph{following}, \emph{yielding})~\cite{munje2025socialnavsub}. This task tests the model's capacity to explicitly classify interaction intents in complex multi-agent environments.

\paragraph{Training and Evaluation}
All reported results are averaged across 10 random seeds and use early stopping~\cite{prechelt2002early} to limit overfitting. 
Model-specific hyperparameter selection, capacity, and complete training protocols are detailed in Appendix~\ref{sec:training_eval_protocol}.

\subsection{RQ1 \& RQ2: Downstream Performance and Data Efficiency}
\label{sec:downstream_efficiency}

\paragraph{Experiment:}
To evaluate downstream performance (RQ1) and data efficiency (RQ2), we pretrain \method on the combined training splits of the datasets using the self-supervised reconstruction objective.
After freezing the encoder, we extract the relevant elements of the latent graph representation (details in the Appendix~\ref{sec:training_eval_protocol}) and train the linear probe using progressively larger fractions of the labels from the training split.
We compare our approach against several unstructured baselines: a fully connected, feed-forward Multi-Layer Perceptron (MLP), an Autoencoder to ablate the graph structure while retaining data compression, and a Random Forest to provide a non-deep-learning baseline.
Note that the Random Forest is omitted for SNS due to the dataset's variable-sized, high-dimensional features. The Autoencoder receives the full temporal edge representation and follows the same pooled pretraining and frozen adaptation protocol across labeled-data fractions.
We report $F_1$-Scores for \textit{competence}, \textit{surprise}, and \textit{intention} for SEAN-T and macro $F_1$-Score over \textit{avoid}, \textit{follow}, and \textit{not consider} for SNS, with per-class results reported in the Appendix~\ref{sec:extended_main_results}.

\paragraph{Results:}

Table~\ref{tbl:r1} shows that the heterogeneous (\method-He) and homogeneous (\method-Ho) variants are competitive with unstructured baselines on SEAN-T and achieve stronger performance on most evaluated settings.
While both exhibit competitive results, \method-Ho yields the highest overall scores in most SEAN-T categories. 
\method-He achieves the highest SNS macro $F_1$ across all labeled-data fractions, with per-class results reported in Appendix~\ref{sec:extended_main_results}.
On SNS, unstructured baselines fail to extract relational signal, plateauing near a macro $F_1$-Score of $0.3$, regardless of how much labeled data is provided.
In contrast, \method-He rapidly acquires task-relevant information, achieving $0.362$ with just 1\% of the data and achieves $0.503$ at 100\%, indicating that the heterogeneous formulation is particularly effective for this relational action classification task. We additionally compare against the same heterogeneous graph architecture trained end-to-end on SNS labels. \method-He achieves a macro $F_1$-Score of $0.503$ at 100\% labeled data, compared with $0.445$ for the supervised model; full results are reported in Appendix~\ref{sec:supervised_heterograph_ablation}.
To assess cross-dataset transfer, we pretrain on one dataset and probe on the other. At 100\% labeled data, \method-He pretrained only on SEAN-T achieves $0.497$ macro $F_1$ on SNS compared with $0.503$ under joint pretraining, while SNS-only pretraining achieves $0.801/0.710/0.729$ on the SEAN-T competence/surprise/intention tasks compared with $0.794/0.709/0.721$ under joint pretraining. \method-Ho also transfers across datasets, although joint pretraining provides larger gains; full results across labeled-data fractions are reported in Appendix~\ref{sec:cross_dataset_transfer}.
% Despite these task-specific variations, \method requires much less labeled data than baseline methods to achieve high performance.
% The relational inductive bias inherent in \method allows the model to rapidly generalize even when restricted to 5\% or 10\% of the labeled downstream data. 
% This advantage is especially notable under extreme low-data regimes; for instance, on the Competence task, \method-Ho achieves an $F_1$-Score of $0.612$ using only 1\% of the labeled data, whereas unstructured baselines like the Random Forest~\cite{tsoi2022sean} or MLP require roughly five to ten times as much data to reach comparable performance. 
Finally, the benefit of \method's relational structure is most pronounced on SNS, where \method-He consistently outperforms the temporal Autoencoder across all labeled-data fractions. On SEAN-T, the Autoencoder remains competitive in several settings, suggesting that explicit relational structure provides a greater benefit for pedestrian action classification than for the perception-prediction tasks.

%%% SEAN-T & SNS RESULTS
\begin{table}[tb!p]
\centering
\caption{
Downstream Performance (RQ1) and Data Efficiency (RQ2) on SEAN-T and SNS.
Results showcase $F_1$-Scores ($\mu \pm \sigma$) across progressively scaled fractions of labeled data.
We compare unstructured baseline methods against our proposed \method variants (\textbf{\method-He}: Heterogeneous, \textbf{\method-Ho}: Homogeneous).
To rigorously evaluate both representation quality and data efficiency, \method utilizes a frozen pretrained encoder paired with a lightweight linear probe.} 
\label{tbl:r1}
\setlength{\tabcolsep}{6pt}
\resizebox{\textwidth}{!}{
\begin{tabular}{ll *{6}{c}} 
\toprule
 & \textbf{Method} & \textbf{1\%} & \textbf{5\%} & \textbf{10\%} & \textbf{25\%} & \textbf{50\%} & \textbf{100\%} \\
\midrule

% --- Competence Section ---
\multirow{5}{*}{\rotatebox[origin=c]{90}{\begin{tabular}{@{}c@{}}\textbf{SEAN-T} \\ \textbf{Competence}\end{tabular}}} 
& MLP   & $0.450 \pm 0.22$ & $0.546 \pm 0.19$ & $0.664 \pm 0.15$ & $0.761 \pm 0.03$ & $0.778 \pm 0.02$ & $0.797 \pm 0.02$ \\
& Autoencoder    & $\mathbf{0.663 \pm 0.08}$ & $0.678 \pm 0.05$ & $0.706 \pm 0.05$ & $0.734 \pm 0.02$ & $0.747 \pm 0.02$ & $0.742 \pm 0.02$ \\
& Random Forest  & $0.557 \pm 0.20$ & $0.641 \pm 0.12$ & $0.680 \pm 0.10$ & $0.704 \pm 0.09$ & $0.715 \pm 0.09$ & $0.718 \pm 0.08$ \\
\cmidrule(lr){2-8}
& \textbf{\method-He (Ours)}& $0.635 \pm 0.10$ & $0.673 \pm 0.07$ & $0.726 \pm 0.04$ & $0.762 \pm 0.03$ & $0.781 \pm 0.03$ & $0.794 \pm 0.02$ \\
& \textbf{\method-Ho (Ours)}  & $0.612 \pm 0.14$ & $\mathbf{0.702 \pm 0.08}$ & $\mathbf{0.738 \pm 0.05}$ & $\mathbf{0.765 \pm 0.04}$ & $\mathbf{0.787 \pm 0.03}$ & $\mathbf{0.799 \pm 0.02}$ \\
\midrule

% --- Surprise Section ---
\multirow{5}{*}{\rotatebox[origin=c]{90}{\begin{tabular}{@{}c@{}}\textbf{SEAN-T} \\ \textbf{Surprise}\end{tabular}}} 
& MLP   & $0.217 \pm 0.22$ & $0.250 \pm 0.14$ & $0.423 \pm 0.27$ & $0.450 \pm 0.21$ & $0.630 \pm 0.23$ & $0.720 \pm 0.09$ \\
& Autoencoder    & $\mathbf{0.548 \pm 0.10}$ & $\mathbf{0.597 \pm 0.12}$ & $\mathbf{0.660 \pm 0.07}$ & $\mathbf{0.674 \pm 0.09}$ & $0.712 \pm 0.10$ & $0.739 \pm 0.06$ \\
& Random Forest  & $0.291 \pm 0.24$ & $0.341 \pm 0.19$ & $0.373 \pm 0.20$ & $0.444 \pm 0.16$ & $0.470 \pm 0.15$ & $0.496 \pm 0.15$ \\
\cmidrule(lr){2-8}
& \textbf{\method-He (Ours)}& $0.455 \pm 0.17$ & $0.481 \pm 0.17$ & $0.526 \pm 0.15$ & $0.603 \pm 0.12$ & $0.655 \pm 0.08$ & $0.709 \pm 0.04$ \\
& \textbf{\method-Ho (Ours)}  & $0.440 \pm 0.23$ & $0.541 \pm 0.21$ & $0.548 \pm 0.20$ & $0.610 \pm 0.16$ & $\mathbf{0.713 \pm 0.06}$ & $\mathbf{0.741 \pm 0.03}$ \\
\midrule

% --- Intention Section ---
\multirow{5}{*}{\rotatebox[origin=c]{90}{\begin{tabular}{@{}c@{}}\textbf{SEAN-T} \\ \textbf{Intention}\end{tabular}}} 
& MLP   & $0.479 \pm 0.20$ & $0.533 \pm 0.13$ & $0.606 \pm 0.06$ & $0.659 \pm 0.03$ & $0.693 \pm 0.03$ & $0.703 \pm 0.01$ \\
& Autoencoder    & $0.569 \pm 0.04$ & $0.622 \pm 0.03$ & $0.659 \pm 0.03$ & $0.665 \pm 0.05$ & $0.661 \pm 0.03$ & $0.664 \pm 0.02$ \\
& Random Forest  & $0.592 \pm 0.18$ & $0.665 \pm 0.10$ & $0.685 \pm 0.09$ & $0.692 \pm 0.09$ & $0.700 \pm 0.08$ & $0.698 \pm 0.09$ \\
\cmidrule(lr){2-8}
& \textbf{\method-He (Ours)}& $0.576 \pm 0.08$ & $0.619 \pm 0.05$ & $0.671 \pm 0.04$ & $0.694 \pm 0.03$ & $0.705 \pm 0.03$ & $0.721 \pm 0.03$ \\
& \textbf{\method-Ho (Ours)}  & $\mathbf{0.603 \pm 0.10}$ & $\mathbf{0.671 \pm 0.07}$ & $\mathbf{0.701 \pm 0.06}$ & $\mathbf{0.715 \pm 0.04}$ & $\mathbf{0.723 \pm 0.03}$ & $\mathbf{0.734 \pm 0.02}$ \\
\midrule

% --- Macro F1 Section (SNS) ---
\multirow{5}{*}{\rotatebox[origin=c]{90}{\begin{tabular}{@{}c@{}}\textbf{SNS} \\ \textbf{Macro $F_1$}\end{tabular}}} 
& MLP    & $0.294 \pm 0.00$ & $0.297 \pm 0.01$ & $0.300 \pm 0.01$ & $0.306 \pm 0.02$ & $0.295 \pm 0.01$ & $0.293 \pm 0.00$ \\
& Autoencoder     & $0.308 \pm 0.02$ & $0.309 \pm 0.02$ & $0.307 \pm 0.02$ & $0.310 \pm 0.02$ & $0.309 \pm 0.02$ & $0.332 \pm 0.01$ \\
% & Random Forest  & --- & --- & --- & --- & --- & --- \\
\cmidrule(lr){2-8}
& \textbf{\method-He (Ours)} & $\mathbf{0.362 \pm 0.06}$ & $\mathbf{0.447 \pm 0.05}$ & $\mathbf{0.470 \pm 0.04}$ & $\mathbf{0.493 \pm 0.03}$ & $\mathbf{0.501 \pm 0.03}$ & $\mathbf{0.503 \pm 0.02}$ \\
& \textbf{\method-Ho (Ours)} & $0.344 \pm 0.05$ & $0.431 \pm 0.05$ & $0.453 \pm 0.04$ & $0.468 \pm 0.03$ & $0.471 \pm 0.02$ & $0.475 \pm 0.02$ \\
\bottomrule
\end{tabular}
}
\end{table}

\subsection{RQ3: Latent Space Semantics}
\label{sec:latent_space_semantics}

\paragraph{Experiment:}
To qualitatively evaluate whether the learned representations capture high-level social semantics without optimizing the encoder with a supervised classification objective (RQ3), we analyze the topology of the latent spaces.
Using the pretrained latent representation from the homogeneous graph encoder reported in Table~\ref{tbl:r1}, we extract frozen representations from the 128-dimensional node-level embeddings for SNS.
To focus on semantic navigation intents, we conduct our experiments on the dominant action labels in the SocialNav-SUB (SNS) dataset, \textit{avoid} and \textit{follow}.
We apply both Principal Component Analysis (PCA) for linear projection and t-Distributed Stochastic Neighbor Embedding (t-SNE) for non-linear dimensionality reduction to map the high-dimensional representations to a 2D space, color-coded by downstream class labels.

\paragraph{Results:}
Latent space projections shown in Figure~\ref{fig:ds_sns} reveal structural self-organization corresponding to downstream human navigation semantics. 
The linear PCA projection (left) shows smooth variation across the latent space, while the non-linear t-SNE projection (right) reveals visually distinct groupings associated with action classes.
Together, these projections provide qualitative evidence that the learned representations contain action relevant structure, despite the encoder not being optimized with a supervised action classification objective.

\section{Limitations and Future Work}
\label{sec:limitations}
%\paragraph{Limitations and Future Work.}
While \method generalizes well, its autoencoder and message passing inherently smooth high-frequency pairwise geometries.
Therefore, these latents excel at node-level tasks like intent prediction, but are less effective on edge-centric tasks (e.g., group detection) that require exact relational geometry.
Additionally, standardizing diverse datasets into unified graphs can discard dataset-specific nuances.
Future work can explore more flexible graph construction pipelines and the use of intermediate, pre-message-passing temporal edge embeddings to bypass the autoencoder bottleneck, preserving the precise geometric information necessary for fine-grained relational tasks.
Our evaluation assumes fixed-duration interaction windows and processed agent tracks, and is limited to curated social navigation datasets. A promising direction for future work is to evaluate \method under online segmentation, noisier perception, and more diverse real-world settings.

\section{Conclusion}
\label{sec:conclusion}
In this work, we introduced \method for self-supervised representation learning in social navigation by bridging continuous physical dynamics and high-level social reasoning.
By modeling scenarios as spatiotemporal graphs and using a self-supervised graph autoencoder, \method injects a strong relational inductive bias. 
Evaluations show that our learned representations are highly reusable and data-efficient, and can encode complex social semantics without directly optimizing the encoder for downstream tasks.
Ultimately, our results provide evidence that structural priors can help offset the scarcity of grounded HRI data, offering a promising direction toward reusable and data-efficient representations for social navigation.

% The acknowledgments are automatically included only in the final and preprint versions of the paper.
\acknowledgments{

A portion of work has taken place in the Learning Agents Research
Group (LARG) at UT Austin. LARG research is supported in part by NSF
(FAIN-2019844, NRT-2125858, OIA-2535195), ONR (N00014-24-1-2550), ARO
(W911NF-17-2-0181, W911NF-23-2-0004, W911NF-25-1-0065), Lockheed
Martin, Lyda Hill, and Good Systems.  Peter Stone serves as the Chief
Scientist of Sony CTC and receives financial compensation for that
role.  The terms of this arrangement have been reviewed and approved
by the University of Texas at Austin in accordance with its policy on
objectivity in research.
Another portion of this work has taken place at the Autonomous Mobile Robotics Laboratory (AMRL) at the Artificial Intelligence Laboratory, The University of Texas at Austin. AMRL research is supported in part by the National Science Foundation (OIA-2535195, CAREER-2046955, OIA-2219236, DGE-2125858, CCF-2319471). 
This work used computing resources at the Texas Advanced Computing Center (TACC) made possible by a generous gift from NVIDIA Robotics. 
Any opinions, findings, and conclusions expressed in this material are those of the authors and do not necessarily reflect the views of the donors or sponsors.
}

\clearpage

%===============================================================================

\bibliography{refs}  % .bib

\newpage
\appendix

\section{Dataset and Graph Statistics}
\label{sec:datasets}

\subsection{Datasets}
We evaluate our method on two datasets providing physical robot-human interaction data: \textbf{SEAN Together} and \textbf{SocialNav-SUB}. Both datasets are converted into the common graph format described in Section~\ref{sec:method}.

\textbf{SEAN Together (SEAN-T).} SEAN Together contains 8-second (5\,Hz) immersive virtual-reality observations of robot navigation behavior paired with subjective human ratings \cite{zhang2023predicting}. Participants rated interactions along three dimensions: perceived competence, clarity of intention, and surprise. We use this to evaluate the extraction of subjective social impressions from physical motion. We organize the SEAN-T dataset by participant and then split it into train, validation, and test subsets corresponding to 80\%, 16\% and 4\% of participants.

\textbf{SocialNav-SUB (SNS).} SocialNav-SUB expands the SCAND dataset~\cite{karnan2022socially} into a structured social reasoning benchmark for navigation scenes~\cite{munje2025socialnavsub}. The dataset contains robot-centric interaction scenarios with relational action labels, such as \emph{avoiding}, \emph{following}, and \emph{yielding to}. We use this dataset to evaluate action classification from physical interaction dynamics. Training uses automatically generated action labels from the SocialNav-SUB heuristic labeling procedure, with the training portion split into 85\% training and 15\% validation, while evaluation uses a separate held-out test set annotated by human labelers.

\begin{table}[h]
\centering
\caption{
Dataset and graph construction details.
The temporal window is fixed within each dataset.
The maximum number of edges follows from the maximum number of nodes under the fully connected graph construction.
}
\label{tab:dataset_graph_sizes}
\setlength{\tabcolsep}{5pt}
\resizebox{\textwidth}{!}{
\begin{tabular}{lcccccc}
\toprule
\textbf{Dataset} & \textbf{\# Scenarios} & \textbf{Sample Freq.} & \textbf{Window Size $T$} & \textbf{Window Duration} & \textbf{Max Nodes $N$} & \textbf{Max Edges} \\
\midrule
SEAN-T & 2969 & 5\,Hz & 40 & 8s & 16 & 240 \\
SocialNav-SUB & 3052 & 8\,Hz & 20 & 2.5s & 31 & 930 \\
\bottomrule
\end{tabular}
}
\end{table}

\FloatBarrier
\clearpage

\section{Implementation Details}
\label{sec:implementation}

\subsection{Graph Construction Details}
\label{sec:stems}

\subsubsection{Shared Graph Representation}

All variants represent each fixed-duration interaction scenario as the spatiotemporal graph $G$ of Eq.~\ref{eq:scenario_graph}, where nodes correspond to observed agents and directed edges correspond to ordered pairwise interactions.
Each edge stores a temporal sequence of relative spatial and motion features, enumerated in Appendix~\ref{sec:edge_features}.
The homogeneous and heterogeneous variants differ only in how entity and relation type information is represented, and in which components are instantiated per dataset (Appendix~\ref{sec:sharing}).
\subsubsection{Edge Feature Channels}
\label{sec:edge_features}

Both variants use the same $d_e = 8$ edge channels, differing only in sequence length.
The first four are the relative $SE(2)$ pose of node $j$ in the body frame of node $i$, and the remaining four are their temporal deltas, $\delta_t u = u_t - u_{t-1}$:
\begin{equation}
\label{eq:edge_channels}
\mathbf{e}_{ij,t} = \bigl[\,\Delta x_{ij,t},\; \Delta y_{ij,t},\; \cos\Delta\theta_{ij,t},\; \sin\Delta\theta_{ij,t},\; \delta_t \Delta x_{ij},\; \delta_t \Delta y_{ij},\; \delta_t \cos\Delta\theta_{ij},\; \delta_t \sin\Delta\theta_{ij} \,\bigr]
\end{equation}
Each delta is set to zero at the first observed timestep of a contiguous track segment, so a pedestrian who leaves and re-enters the window contributes no spurious jump across the gap.
Because headings enter only through $(\cos, \sin)$ pairs, the encoding has no wrap-around discontinuity.
The deltas are per-timestep differences rather than rates, so their scale depends on the sampling frequency; features are standard-scaled before training, which removes the resulting offset between datasets but not the difference in what one timestep represents physically.

\subsubsection{Homogeneous Graph}
\label{sec:homo-graph-construction}

\textbf{Unified Representation:}
The homogeneous model processes both SEAN Together and SocialNav-SUB (SNS) as single-type graphs, while robot and human identity is included only as a 2-class input node embedding. All data is unified to a 40-timestep window, with SNS interpolated from 20 to 40 steps, and projected into a shared 592-dimensional node feature space.

\textbf{Feature Sets:}
\begin{itemize}
    \item \textbf{Node Features:} SEAN robot nodes contain 512D map features and 80D goal trajectories. For SNS robot nodes and all human nodes across both datasets, these 592 dimensions are zero-initialized. A bias-free linear projection is used so that zeroed features contribute only through the learnable type embeddings.
    \item \textbf{Edge Features:} 8-dimensional relative SE(2) poses and temporal deltas (Appendix~\ref{sec:edge_features}).
\end{itemize}

\textbf{Processing:}
A \textit{Temporal Edge Transformer} encodes the 40-timestep edge attributes into fixed-size embeddings. Message passing is performed using a homogeneous \textit{TransformerConv} layer that operates on the entire graph simultaneously. An edge-level padding mask is derived from the \textit{valid mask} to ensure the model ignores invalid timesteps during temporal pooling.

\subsubsection{Heterogeneous Graph}
\label{sec:hetero-graph-construction}

\textbf{Multi-Type Topology:}
The heterogeneous model utilizes a \textit{HeteroData} structure with two node types (\textit{robot}, \textit{bystander}) and three directed edge types: (\textit{robot, to, bystander}), (\textit{bystander, to, robot}), and (\textit{bystander, to, bystander}). Unlike the homogeneous case, SEAN (40 steps) and SNS (20 steps) retain their native temporal resolutions, and batches are restricted to a single source to prevent temporal mismatch. For SEAN-only heterogeneous experiments, we additionally retain the SEAN-specific \textit{follower} node, adding directed edges between the follower and both robot and bystander nodes. This preserves the full interaction structure available in SEAN while keeping the combined heterogeneous model restricted to node and edge types shared across datasets.

\textbf{Feature Sets:}
\begin{itemize}
    \item \textbf{SEAN Nodes:} The robot uses 592D features (map + goal). Bystanders use a learnable 8D embedding concatenated with an 8-dimensional positional feature derived from the mean-pooled relative pose of the \textit{robot-to-bystander} edge.
    \item \textbf{SNS Nodes:} Both robot and bystander nodes use a learnable 8D embedding concatenated with an 8-dimensional positional feature derived from the mean-pooled relative pose of the given node's incident \textit{robot-to-bystander} edges.
    \item \textbf{Edge Features:} 8-dimensional SE(2) relative pose and dynamics, matching the dimensions of the homogeneous case but varying in sequence length.
\end{itemize}

\begin{table}[t]
\centering
\small
\caption{Node features by architecture and dataset.
All configurations use 8D edge features comprising relative position,
relative heading (sine and cosine), and their temporal differences.
Both architectures use reconstruction pretraining followed by
linear probing with a frozen encoder.}
\label{tab:features_training}
\setlength{\tabcolsep}{4pt}
\renewcommand{\arraystretch}{1.15}
\begin{tabularx}{\textwidth}{
    @{}ll
    >{\raggedright\arraybackslash}X
    >{\raggedright\arraybackslash}X@{}}
\toprule
\textbf{Model} & \textbf{Dataset} &
\textbf{Robot features} & \textbf{Human features} \\
\midrule

\method-He & SEAN-T &
592D map and goal features &
8D learned embedding + 8D pooled edge features \\
\addlinespace

\method-He & SNS &
8D learned embedding + 8D pooled edge features &
8D learned embedding + 8D pooled edge features \\
\midrule

\method-Ho & SEAN-T &
592D map and goal features + type embedding &
592D zero vector + type embedding \\
\addlinespace

\method-Ho & SNS &
592D zero vector + type embedding &
592D zero vector + type embedding \\

\bottomrule
\end{tabularx}
\end{table}

\subsection{Graph Encoder Details}

\subsubsection{Per-Dataset and Shared Components}
\label{sec:sharing}

Because $d_v^{(k)}$ and $T_k$ differ across datasets, components whose input or output dimensionality depends on these quantities are instantiated per dataset, while the relational message-passing core and latent heads are shared.
Table~\ref{tab:sharing} gives the split, which differs between the two variants and is the main structural distinction between them: the homogeneous variant unifies the datasets at conversion time and shares every component, while the heterogeneous variant keeps them separate through the input and output adapters and shares the relational core.
In the heterogeneous variant this means four node feature projections, one per dataset and node type, three learned type embeddings, two temporal edge encoders, and two edge decoders.

\begin{table}[h]
\centering
\caption{Components shared across datasets in each variant.}
\label{tab:sharing}
\setlength{\tabcolsep}{6pt}
\begin{tabular}{lcc}
\toprule
\textbf{Component} & \textbf{Homogeneous} & \textbf{Heterogeneous} \\
\midrule
Node feature projection        & shared            & per dataset and node type \\
Temporal edge encoder          & shared ($T{=}40$) & per dataset ($40$, $20$) \\
Message-passing layers         & shared            & shared \\
Edge-type embedding            & not used          & shared \\
Latent heads $f_\mu, f_\sigma$ & shared            & shared \\
Edge decoder                   & shared ($T{=}40$) & per dataset \\
Node decoder                   & \multicolumn{2}{c}{SEAN-T robot node only} \\
\bottomrule
\end{tabular}
\end{table}

\subsubsection{Feature Preprocessing}
Node attributes are first projected into a shared hidden dimension $d_{hidden}$ via type-specific Multilayer Perceptrons (MLPs):
\begin{equation}
\mathbf{h}_i^{(0)} = \text{MLP}_{\tau(i)}(\mathbf{x}_i)
\end{equation}
where $\tau(i)$ denotes the node type. To encode temporal edge sequences across datasets, temporal edge sequences $\mathbf{e}_{ij}$ are passed through a Transformer edge encoder with learned positional embeddings. This compresses the full temporal sequence into a fixed-dimensional relation embedding $\tilde{\mathbf{e}}_{ij} \in \mathbb{R}^{d_{hidden}}$:

\begin{equation}
\tilde{\mathbf{e}}_{ij} = \text{Enc}_{\text{edge}}(\mathbf{e}_{ij})
\end{equation}

\subsubsection{Message Passing}
\label{sec:message-passing}

\textbf{Heterogeneous Message Passing}

The graph encoder performs message passing over the complete interaction graph. In our heterogeneous variant, we utilize a HEAT-style update~\cite{mo2021heat}. Let $\mathbf{h}_j^{(k)}$ and $\mathbf{h}_i^{(k)}$ denote the source and destination node embeddings at layer $k$, and $\mathbf{t}_{ij}$ denote a learned embedding for the specific edge type. We construct an edge-conditioned interaction representation:
\begin{equation}
\mathbf{u}_{ij}^{(k)} = \left[ \mathbf{h}_j^{(k)} \parallel \mathbf{h}_i^{(k)} \parallel \tilde{\mathbf{e}}_{ij}^{(k)} \parallel \mathbf{t}_{ij} \right]
\end{equation}
This representation encapsulates the full context of the interaction. During message passing, the network must determine both the content of the information being transmitted and the relative importance of that information. Therefore, we use $\mathbf{u}_{ij}^{(k)}$ to independently parameterize both the raw attention score $s_{ij}^{(k)}$ and the message content $\mathbf{m}_{ij}^{(k)}$ via two learnable functions:
\begin{equation}
s_{ij}^{(k)} = f_{\mathrm{attn}}\left(\mathbf{u}_{ij}^{(k)}\right), \qquad \mathbf{m}_{ij}^{(k)} = f_{\mathrm{msg}}\left(\mathbf{u}_{ij}^{(k)}\right)
\end{equation}
Formally, the destination-wise softmax normalizes the raw attention scores across all incoming edges from the neighborhood $\mathcal{N}(i)$:
\begin{equation}
\alpha_{ij}^{(k)} = \frac{\exp\left(s_{ij}^{(k)}\right)}{\sum_{m \in \mathcal{N}(i)} \exp\left(s_{im}^{(k)}\right)}
\end{equation}
These normalized weights are used to compute the aggregated neighborhood message $\bar{\mathbf{m}}_i^{(k)}$:
\begin{equation}
\bar{\mathbf{m}}_i^{(k)} = \sum_{j \in \mathcal{N}(i)} \alpha_{ij}^{(k)} \mathbf{m}_{ij}^{(k)}
\end{equation}
The node state is updated via a residual connection and Layer Normalization ($\mathrm{LN}$) to maintain numerical stability during training:
\begin{equation}
\mathbf{h}_i^{(k+1)} = \mathrm{LN}\!\left( \mathbf{h}_i^{(k)} + f^{\mathrm{upd}}\left( [\mathbf{h}_i^{(k)} \parallel \bar{\mathbf{m}}_i^{(k)}] \right) \right)
\end{equation}
To enhance our edges' influence in the GNN, and for future edge-level downstream task integration, we additionally update edge embeddings at each layer following the Group GNN convention~\cite{thompson2025social}. The message content $\mathbf{m}_{ij}^{(k)}$, which was already computed for node aggregation previously, is reused as a residual update to the edge embedding:
\begin{equation}
\tilde{\mathbf{e}}_{ij}^{(k+1)} = \mathrm{LN}_{\mathrm{edge}}\!\left( \tilde{\mathbf{e}}_{ij}^{(k)} + \mathbf{m}_{ij}^{(k)} \right)
\end{equation}
We perform a similar computation as the node update equation, except we omit a separate update function for $\mathbf{m}_{ij}^{(k)}$ since it has already been transformed by an MLP, making an additional transformation redundant.

\textbf{Homogeneous Message Passing}

For the homogeneous graph encoder, message passing does not use the HEAT-style update above.
It instead applies a shared edge-conditioned attention convolution~\cite{shi2021maskedlabelpredictionunified} in which the encoded temporal edge features $\tilde{\mathbf{e}}_{ij}$ enter the key and value terms, with no edge-type embedding.

\begin{figure}[p!tb]
\centering
\resizebox{\textwidth}{!}{
\begin{tikzpicture}[
    node distance=1.0cm and 1.2cm,
    box/.style={rectangle, draw=black, thick, fill=white, minimum width=3.4cm, minimum height=1.3cm, align=center, font=\small},
    header/.style={font=\fontseries{b}\selectfont\small, align=center, text width=3.5cm},
    line/.style={draw, -{Stealth[scale=1.2]}, thick},
    dashline/.style={draw, -{Stealth[scale=1.2]}, thick, dashed}
]

    \node[header] (h1) {RAW INPUTS};
    
    \node[box, below=0.4cm of h1] (robot_in) {
        \textbf{Robot Node} \\ 592D \\ 
        \scriptsize 512 occ grid + 80 goal
    };
    
    \node[box, below=1.2cm of robot_in] (human_in) {
        \textbf{Human Nodes} \\ 16D \\
        \scriptsize 8D learned embedding \\
        \scriptsize + 8D positional feature
    };
    
\node[box, below=1.2cm of human_in] (edge_in) {
    \textbf{Edge Features} \\
    \scriptsize $T_k \times 8D$ ($T{=}40$ SEAN-T, $T{=}20$ SNS) \\
    \scriptsize (rel. $SE(2)$ pose + temporal deltas)
};

    \node[header, right=of h1] (h2) {PREPROCESSING};
    
    \node[box, right=of robot_in] (robot_pre) {
        \textbf{MLP} \\ $\rightarrow$ 256D
    };
    
    \node[box, right=of human_in] (human_pre) {
        \textbf{MLP + Pos} \\ $\rightarrow$ 256D \\
        \scriptsize 8D learned embedding \\
        \scriptsize + 8D positional embedding
    };
    
    \node[box, right=of edge_in] (edge_pre) {
        \textbf{Transformer} \\ \scriptsize $3 \text{ layers}, 4 \text{ heads} \rightarrow 256D$
    };

    \node[box, rectangle, minimum width=3.6cm, minimum height=5.8cm, right=1.1cm of human_pre] (encoder) {};
    \node[header, above=0.2cm of encoder.north] (h3) {ENCODER};
    
    % Graph Visuals (Centered using encoder.center)
    \node[circle, draw, inner sep=2.5pt, font=\scriptsize] (r_node) at ($(encoder.center)+(-0.5, 1.2)$) {R};
    \node[circle, draw, inner sep=2.5pt, font=\scriptsize] (h_node1) at ($(encoder.center)+(-0.3, -0.3)$) {H};
    \node[circle, draw, inner sep=2.5pt, font=\scriptsize] (h_node2) at ($(encoder.center)+(0.6, 0.4)$) {H};
    
    \node[above=0.05cm of r_node, font=\tiny] {Robot};
    \node[below=0.05cm of h_node1, font=\tiny] {Human};
    
    \path[draw, thick] (r_node) -- node[left, font=\tiny, inner sep=2pt] {RH edge} (h_node1);
    \path[draw, thick] (h_node1) -- node[below right, font=\tiny, inner sep=1pt] {H H edge} (h_node2);
    \path[draw, thick] (r_node) -- (h_node2);
    
    % Bottom Equations Layer (Rule width adjusted from 5.2cm to 3.2cm to match new box width)
    \node[align=center, font=\scriptsize, anchor=south, yshift=0.2cm, text width=3.4cm] (enc_equations) at (encoder.south) {
        \rule{3.2cm}{0.4pt} \\[0.15cm]
        \textbf{2 MP Layers} \\
        \textbf{256D Hidden Dim} \\[0.15cm]
        $u_{ij}=[h_{j}||h_{i}||\overline{e}_{ij}||t_{ij}]$ \\[0.1cm]
        $\alpha_{ij}=\text{softmax}_{j}(f_{\text{attn}}(u_{ij}))$
    };

    \node[box, rectangle, minimum width=4.2cm, minimum height=4.6cm, right=1.2cm of encoder] (latent) {
        \textbf{\small LATENT SPACE} \\[0.15cm]
        \rule{3.8cm}{0.4pt} \\[0.15cm]
        \scriptsize $\mu_{i}$, $\log \sigma_{i}^{2}$ shared linear heads \\ \vspace{0.15cm}
        \scriptsize $z_{i}=\mu_{i}+\sigma_{i}\odot\epsilon$ \\
        \scriptsize 128D per node \\ \vspace{0.15cm}
        \scriptsize $D_{\text{KL}}(q_{\phi}||\mathcal{N}(0,I))$
    };
    \node[header, above=0.2cm of latent.north] (h4) {LATENT SPACE};

    \node[box, right=1.2cm of latent.north east, anchor=north west, minimum width=3.6cm] (edge_dec) {
        \textbf{Edge MLP} \\
        \scriptsize $[z_{i}||z_{j}||t_{ij}]$ \\
        \scriptsize $\rightarrow$ recon. edges
    };
    \node[header, above=0.2cm of edge_dec.north] (h5) {DECODER};
    
    \node[box, right=1.2cm of latent.south east, anchor=south west, minimum width=3.6cm] (robot_dec) {
        \textbf{Robot MLP} \\
        \scriptsize $Z_{\text{robot}}$ \\
        \scriptsize $\rightarrow$ recon. robot feat \\
        \scriptsize (SEAN only)
    };
    
    % Downstream Task: Connected via dashed line from Latent Box
    \node[box, below=1.2cm of latent, minimum width=4.2cm, dashed] (lin_probe) {
        \textbf{Linear Probe (SLA)} \\
        \scriptsize $\downarrow$ Frozen z
    };

    % Global Loss Statement (Final optimization objective boundary)
    \node[anchor=north east, font=\small] (loss) at ($(robot_dec.south east)+(0, -0.5)$) {
        $\mathcal{L}=\mathcal{L}_{\text{recon}}+\beta\cdot\mathcal{L}_{\text{KL}}$
    };

    \path[line] (robot_in) -- (robot_pre);
    \path[line] (human_in) -- (human_pre);
    \path[line] (edge_in) -- (edge_pre);
    
    \path[line] (robot_pre.east) -- ($(encoder.west)!(robot_pre.east)!(encoder.west)$);
    \path[line] (human_pre.east) -- (encoder.west);
    \path[line] (edge_pre.east) -- ($(encoder.west)!(edge_pre.east)!(encoder.west)$);
    
    \path[line] (encoder) -- (latent);
    
    % Paths going into Decoder Heads
    \path[line] ($(latent.east)!(edge_dec.west)!(latent.east)$) -- (edge_dec.west);
    \path[line] ($(latent.east)!(robot_dec.west)!(latent.east)$) -- (robot_dec.west);
    
    % Isolated Downstream Pathway (KL path / frozen routing)
    \path[dashline] (latent.south) -- (lin_probe.north);

\end{tikzpicture}
} 
\caption{\textbf{STARS Heterogeneous architecture overview.} \textit{Raw inputs} (robot occupancy + goal, learned human embeddings, 8D temporal edge features) are projected to a 256D shared hidden space via type-specific \textit{preprocessing} (MLPs for nodes, a Transformer for edges). The \textit{encoder} performs HEAT-style message passing (MP) over the fully connected heterogeneous graph (2 layers), with attention and messages parameterized over the concatenated edge-conditioned representation $\mathbf{u}_{ij}$. The resulting 256D node states are projected through shared linear heads to per-node Gaussian posteriors and sampled to 128D latents $\mathbf{z}_i$ via the reparameterization trick. The \textit{decoder} reconstructs edge features (and robot features on SEAN) from latent pairs, trained jointly with a KL divergence term against $\mathcal{N}(0, I)$. Pretrained encoders are frozen for downstream linear probing (SLA).}
\label{fig:heter_architecture}
\end{figure}
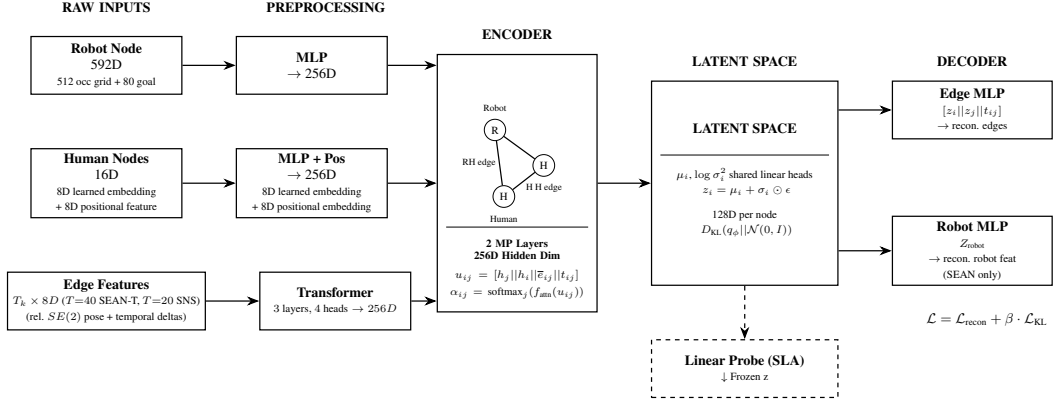

\FloatBarrier
\clearpage
\section{Training and Evaluation Protocol}
\label{sec:training_eval_protocol}

All experiments were implemented in PyTorch~\cite{paszke2019pytorch} using mixed precision.

We applied standard feature scaling and fast early stopping~\cite{prechelt2002early} with patience $20$.
All reported results are averaged across 10 random seeds.
Models were optimized with AdamW using decoupled weight decay~\cite{loshchilov2017decoupled}; model-specific learning rates and selected hyperparameters are reported in Appendix~\ref{sec:hyperparameter_search}.

\subsection{Pretraining Protocol}
\label{sec:pretraining_protocol}

Models were first pretrained using the ELBO self-supervised VAE objective
described in Sec.~\ref{sec:method}. Graphs were processed independently
within each minibatch, allowing each interaction scenario to contain a
variable number of human nodes and edges. Optimization details and selected
hyperparameters are provided in Appendix~\ref{sec:hyperparameter_search}.

\subsection{Downstream Adaptation Protocol}
\label{sec:downstream_protocol}

After pretraining, encoder weights were frozen and reused for downstream
prediction tasks. For supervised linear adaptation (SLA), latent graph
representations were extracted from the pretrained encoder and passed into
lightweight linear prediction heads. Linear probes were trained
independently for each downstream task while keeping the graph encoder
fixed.

The probe input follows Eq.~\ref{eq:readouts}: for the graph-level SEAN-T tasks it concatenates the robot latent with the mean of the human latents, giving one $2 d_z = 256$ dimensional vector per scenario, and for the node-level SNS task it concatenates the robot latent with that pedestrian's own latent, giving one 256-dimensional vector per labeled pedestrian.
The three independent binary SEAN-T heads therefore hold $3 \times (256 + 1) = 771$ parameters and the single five-way SNS head $5 \times (256 + 1) = 1285$, matching Table~\ref{tab:model_capacity}.
The SNS classifier retains all five action categories, but we report macro $F_1$ over \emph{avoid}, \emph{follow}, and \emph{not consider}. 
We exclude \emph{overtake} and \emph{yield} from the aggregate metric because their limited labeled support does not permit stable per-class evaluation.

\subsection{Hyperparameter Search}
\label{sec:hyperparameter_search}

We tuned hyperparameters only for the unified SEAN+SNS models. Hyperparameter search was performed with Optuna using a TPE sampler and median pruning. Each trial first trained a unified GNN-VAE, then froze the encoder and trained an SNS linear probe. The Optuna objective was SNS probe validation loss using cross-entropy.

For the homogeneous graph model, the search selected a batch size of $32$, dropout of $0.1$, weight decay of $10^{-4}$, adaptive maximum mixing coefficient of $0.75$, $\beta_{\max}=0.1$, hidden dimension $256$, latent dimension $128$, three Transformer layers, and learning rate $10^{-3}$.

For the heterogeneous graph model, we ran three independent Optuna searches with $\beta_{\max}$ fixed at $0.1$, $0.5$, and $0.9$, optimizing the remaining hyperparameters within each search. The $\beta_{\max}=0.9$ search yielded the best downstream validation performance. The selected heterogeneous configuration used hidden dimension $256$, latent dimension $128$ per node, batch size $32$, two HEAT-style message-passing layers, and a temporal edge Transformer with depth $3$ and $4$ attention heads. Training used a VAE learning rate of $1.156 \times 10^{-4}$, KL warmup of $25$ epochs, dropout $0.2$, adaptive sampling warmup of $20$ epochs, adaptive ramp-up of $100$ epochs, and maximum mixing coefficient $0.75$.

All final unified models used AdamW optimization, VAE weight decay $10^{-4}$, probe weight decay $10^{-4}$, gradient clipping at $20.0$, VAE early stopping patience $20$, and probe early stopping patience $20$. The unified search used SEAN graphs from \texttt{combined\_v1.h5}, SNS graphs from \texttt{social\_nav\_v3.h5}, and the fixed scene split \texttt{split\_indices\_final.pt}. SNS node features were dropped, so SNS encodings used learned node-type embeddings and edge geometry.

For the SEAN-only baseline, we did not run a separate Optuna search. Instead, the SEAN-only model reused the selected unified hyperparameters wherever applicable, including hidden dimension, latent dimension, learning rate, KL weight, KL warmup, dropout, batch size, weight decay, and gradient clipping. Unified-only adaptive SNS sampling parameters were omitted for SEAN-only training.

\subsection{Sampling and Class Imbalance}
\label{sec:sampling}

We used stratified and adaptive sampling during unified GNN training for both heterogeneous and homogeneous models to address the large SNS class imbalance, where approximately $82\%$ of training examples correspond to the \textit{not consider} class. Although the VAE pretraining objective itself is unsupervised, the adaptive sampler uses downstream SNS label statistics to improve the class balance of batches used during unified training. Empirically, this sampling strategy improved downstream $F_1$ scores for both SEAN and SNS tasks, with particularly noticeable gains for the homogeneous graph model.

\newpage
\section{Compute and Reproducibility Details}
\label{sec:compute_details}

To support reproducibility and clarify the computational requirements of \method, we report dataset sizes, graph construction details, model capacity, training runtimes, inference runtimes, and hardware details for all experiments.

\subsection{Hardware and Software}
\label{sec:hardware}

\paragraph{Hardware.}
All experiments were run primarily on the Lonestar6 and Stampede3 systems at the Texas Advanced Computing Center (TACC). Unless otherwise stated, models were trained on a single NVIDIA A100 PCIe GPU with 40 GB of GPU memory. Experiments were executed in an Ubuntu 22.04 containerized environment using Python 3.10, PyTorch 2.0.1, and CUDA-enabled GPU execution.

\subsection{Model Capacity}
\label{sec:model_capacity}
Table~\ref{tab:model_capacity} reports the number of trainable parameters for \method.
For \method, we report the capacity of the pretrained encoder-decoder separately from the lightweight downstream prediction head.
% For baseline models, we report the number of trainable parameters or, for non-neural baselines, the relevant model complexity settings.
\begin{table}[h]
  \centering
  \caption{
  Model capacity comparison.
  For \method, parameter counts are separated into the pretrained representation model and the downstream task-specific prediction head.
  }
  \label{tab:model_capacity}
  \setlength{\tabcolsep}{5pt}
  \resizebox{\textwidth}{!}{
  \begin{tabular}{lcccc} 
  \toprule
  \textbf{Model / Component} & \textbf{Stage} & \textbf{Trainable Params.} & \textbf{Frozen Params.} & \textbf{Notes} \\
  \midrule
  \method-He Encoder/Decoder & Pretraining & 6{,}420{,}810 & N/A & Heterogeneous graph autoencoder \\
  \method-Ho Encoder/Decoder & Pretraining & 2{,}936{,}848 & N/A & Homogeneous graph autoencoder \\
  Linear Prediction Head & Downstream adaptation & 1{,}285 (SNS), 771 (SEAN) & 6{,}420{,}810 & Used with frozen \method encoder \\
  \bottomrule
  \end{tabular}
  }
  \end{table}
\subsection{Runtime and Memory}

\paragraph{Training and inference runtime.}
Table~\ref{tab:runtime_memory} reports the computational cost of each experimental pipeline.
Pretraining time refers to the wall-clock time required to train the graph autoencoder.
Downstream training time refers to fitting the task-specific prediction head on top of the frozen encoder.
Inference runtime is reported per interaction scenario and includes representation extraction and prediction when applicable.
Peak GPU memory is measured during training unless otherwise stated.

 \begin{table}[h]
  \centering
  \caption{
  Runtime and memory usage for each experimental pipeline.
  For \method-based pipelines, pretraining is performed once and the frozen encoder is reused for downstream tasks.
  Downstream training time refers to fitting the task-specific prediction head.
  Inference runtime is reported per interaction scenario.
  }
  \label{tab:runtime_memory}
  \setlength{\tabcolsep}{4pt}
  \resizebox{\textwidth}{!}{
  \begin{tabular}{lccccc}
  \toprule
  \textbf{Model / Pipeline} & \textbf{Dataset / Task} & \textbf{Pretraining Time} & \textbf{Downstream Training Time} & \textbf{Inference Time / Scenario} & \textbf{Peak
  GPU Memory} \\
  \midrule  
  \method-He + Linear Head & SEAN-T + SNS & $\sim$2.5 hrs / run & $<$1 sec & $\sim$12 ms & $\sim$13 GB \\
  \method-Ho + Linear Head & SEAN-T + SNS & $\sim$3 hrs / run & $\sim$ 12 sec  & $\sim$30 ms & $\sim$17 GB \\
  \end{tabular}
  }
  \end{table}

\FloatBarrier
\clearpage
\newpage
\section{Additional Results and Ablations}
\label{sec:additional_results}

\subsection{Extended Main Results}
\label{sec:extended_main_results}

Table~\ref{tbl:r2} provides the full downstream results for \method across all supervised data fractions. We report SEAN-T performance for the three subjective social-impression tasks and SNS performance using per-class $F_1$ scores together with macro-$F_1$. Across both datasets, performance generally improves as the amount of labeled data available for supervised linear adaptation increases. The heterogeneous and homogeneous variants show similar trends on SEAN-T, while the heterogeneous model achieves higher SNS macro-$F_1$ at larger label fractions.

\begin{table*}[t]
\centering
\small
\caption{Extended \method results on SEAN-T and SNS across labeled-data fractions.}
\label{tbl:r2}
\resizebox{\textwidth}{!}{
\begin{tabular}{llcccccc}
\toprule
\textbf{Task} & \textbf{Method} & \textbf{1\%} & \textbf{5\%} & \textbf{10\%} & \textbf{25\%} & \textbf{50\%} & \textbf{100\%} \\
\midrule
\multirow{2}{*}{Competence} 
& \method-He (Ours) & $0.635 \pm 0.10$ & $0.673 \pm 0.07$ & $0.726 \pm 0.04$ & $0.762 \pm 0.03$ & $0.781 \pm 0.03$ & $0.794 \pm 0.02$ \\
& \method-Ho (Ours) & $0.612 \pm 0.14$ & $0.702 \pm 0.08$ & $0.738 \pm 0.05$ & $0.765 \pm 0.04$ & $0.787 \pm 0.03$ & $0.799 \pm 0.02$ \\
\midrule
\multirow{2}{*}{Surprise} 
& \method-He (Ours) & $0.455 \pm 0.17$ & $0.481 \pm 0.17$ & $0.526 \pm 0.15$ & $0.603 \pm 0.12$ & $0.655 \pm 0.08$ & $0.709 \pm 0.04$ \\
& \method-Ho (Ours) & $0.440 \pm 0.23$ & $0.541 \pm 0.21$ & $0.548 \pm 0.20$ & $0.610 \pm 0.16$ & $0.713 \pm 0.06$ & $0.741 \pm 0.03$ \\
\midrule
\multirow{2}{*}{Intention} 
& \method-He (Ours) & $0.576 \pm 0.08$ & $0.619 \pm 0.05$ & $0.671 \pm 0.04$ & $0.694 \pm 0.03$ & $0.705 \pm 0.03$ & $0.721 \pm 0.03$ \\
& \method-Ho (Ours) & $0.603 \pm 0.10$ & $0.671 \pm 0.07$ & $0.701 \pm 0.06$ & $0.715 \pm 0.04$ & $0.723 \pm 0.03$ & $0.734 \pm 0.02$ \\
\midrule
\multirow{2}{*}{Avoid} 
& \method-He (Ours) & $0.098 \pm 0.09$ & $0.173 \pm 0.11$ & $0.190 \pm 0.10$ & $0.219 \pm 0.10$ & $0.230 \pm 0.09$ & $0.242 \pm 0.08$ \\
& \method-Ho (Ours) & $0.126 \pm 0.09$ & $0.234 \pm 0.06$ & $0.252 \pm 0.06$ & $0.259 \pm 0.06$ & $0.257 \pm 0.05$ & $0.267 \pm 0.05$ \\
\midrule
\multirow{2}{*}{Follow} 
& \method-He (Ours) & $0.155 \pm 0.15$ & $0.327 \pm 0.12$ & $0.374 \pm 0.08$ & $0.404 \pm 0.04$ & $0.412 \pm 0.03$ & $0.405 \pm 0.03$ \\
& \method-Ho (Ours) & $0.132 \pm 0.13$ & $0.289 \pm 0.12$ & $0.343 \pm 0.07$ & $0.374 \pm 0.04$ & $0.385 \pm 0.04$ & $0.396 \pm 0.02$ \\
\midrule
\multirow{2}{*}{Not Consider} 
& \method-He (Ours) & $0.834 \pm 0.06$ & $0.840 \pm 0.04$ & $0.846 \pm 0.03$ & $0.855 \pm 0.02$ & $0.861 \pm 0.02$ & $0.862 \pm 0.02$ \\
& \method-Ho (Ours) & $0.773 \pm 0.11$ & $0.768 \pm 0.05$ & $0.765 \pm 0.04$ & $0.771 \pm 0.03$ & $0.770 \pm 0.04$ & $0.763 \pm 0.04$ \\
\midrule
\multirow{2}{*}{Macro $F_1$} 
& \method-He (Ours) & $0.362 \pm 0.06$ & $0.447 \pm 0.05$ & $0.470 \pm 0.04$ & $0.493 \pm 0.03$ & $0.501 \pm 0.03$ & $0.503 \pm 0.02$ \\
& \method-Ho (Ours) & $0.344 \pm 0.05$ & $0.431 \pm 0.05$ & $0.453 \pm 0.04$ & $0.468 \pm 0.03$ & $0.471 \pm 0.02$ & $0.475 \pm 0.02$ \\
\bottomrule
\end{tabular}
}
\end{table*}

\subsection{Pretraining Data Ablations}
\label{sec:pretraining_data_ablations}
\paragraph{SEAN-only Pretraining.}
We compare the jointly pretrained SEAN-T+SNS model against a SEAN-T only model to assess the effect of adding cross-dataset interaction data on SEAN-T downstream performance.
The SEAN-T only model remains competitive with joint pretraining, particularly at larger labeled-data fractions.
At 100\% labeled data, it achieves $0.792/0.716/0.721$ on competence, surprise, and intention, compared with $0.794/0.709/0.721$ under joint pretraining.
These results indicate that adding SNS data does not substantially alter SEAN-T downstream performance.
% We first compare the combined SNS+SEAN pretrained model against a SEAN-only pretrained model to test whether adding cross-dataset interaction data improves SEAN-specific downstream performance. The SEAN-only model performs competitively, especially at higher supervised data fractions, but remains below the combined model across the main SEAN metrics. This suggests that pretraining on additional heterogeneous interaction data doesn't degrade the quality of the learned social representation.

\begin{table*}[t]
\centering
\small
\caption{SEAN-T downstream performance after SEAN-T-only pretraining. Results report $F_1$-Score ($\mu \pm \sigma$) across labeled-data fractions.}
\label{tab:sean_only_model}
\resizebox{\textwidth}{!}{
\begin{tabular}{llcccccc}
\toprule
\textbf{Task} & \textbf{Model} & \textbf{1\%} & \textbf{5\%} & \textbf{10\%} & \textbf{25\%} & \textbf{50\%} & \textbf{100\%} \\
\midrule
Competence & \method-He & $0.568 \pm 0.10$ & $0.636 \pm 0.07$ & $0.686 \pm 0.06$ & $0.733 \pm 0.03$ & $0.773 \pm 0.03$ & $0.792 \pm 0.02$ \\
Surprise & \method-He & $0.455 \pm 0.14$ & $0.453 \pm 0.16$ & $0.480 \pm 0.22$ & $0.55 \pm 0.15$ & $0.653 \pm 0.09$ & $0.716 \pm 0.05$ \\
Intention & \method-He & $0.528 \pm 0.09$ & $0.601 \pm 0.05$ & $0.644 \pm 0.05$ & $0.683 \pm 0.03$ & $0.705 \pm 0.03$ & $0.721 \pm 0.02$ \\
\bottomrule
\end{tabular}
}
\end{table*}

\subsection{Cross-Dataset Transfer}
\label{sec:cross_dataset_transfer}

To evaluate whether \method learns representations that transfer across
datasets, we pretrain on one dataset and evaluate the frozen encoder on
the other using the same linear-probing protocol as the main
experiments. We compare single-dataset pretraining against joint
SEAN-T+SNS pretraining.
\begin{table*}[t]
\centering
\small
\caption{Cross-dataset transfer to SNS. Models are pretrained either
on SEAN-T only or jointly on SEAN-T and SNS and evaluated on SNS using
progressively larger fractions of labeled training data. Results report
macro $F_1$-Score ($\mu \pm \sigma$).}
\label{tab:cross_dataset_sns}
\setlength{\tabcolsep}{4pt}
\resizebox{\textwidth}{!}{
\begin{tabular}{llcccccc}
\toprule
\textbf{Model} & \textbf{Pretraining} &
\textbf{1\%} & \textbf{5\%} & \textbf{10\%} &
\textbf{25\%} & \textbf{50\%} & \textbf{100\%} \\
\midrule
\method-He & SEAN-T only
& $0.355 \pm 0.06$ & $0.442 \pm 0.05$ & $0.463 \pm 0.04$ & $0.489 \pm 0.03$ & $0.493 \pm 0.03$
& $0.497 \pm 0.03$ \\
\method-He & SEAN-T+SNS
& $0.362 \pm 0.06$
& $0.447 \pm 0.05$
& $0.470 \pm 0.04$
& $0.493 \pm 0.03$
& $0.501 \pm 0.03$
& $0.503 \pm 0.02$ \\
\midrule
\method-Ho & SEAN-T only
& $0.274 \pm 0.05$
& $0.332 \pm 0.04$
& $0.351 \pm 0.03$
& $0.365 \pm 0.03$
& $0.407 \pm 0.02$
& $0.419 \pm 0.01$ \\
\method-Ho & SEAN-T+SNS
& $0.344 \pm 0.05$
& $0.431 \pm 0.05$
& $0.453 \pm 0.04$
& $0.468 \pm 0.03$
& $0.471 \pm 0.02$
& $0.475 \pm 0.02$ \\
\bottomrule
\end{tabular}
% \begin{tabular}{llcccccc}
% \toprule
% \textbf{Model} & \textbf{Pretraining} &
% \textbf{1\%} & \textbf{5\%} & \textbf{10\%} &
% \textbf{25\%} & \textbf{50\%} & \textbf{100\%} \\
% \midrule
% \method-He & SEAN-T only
% & $0.355 \pm 0.058$ & $0.442 \pm 0.054$ & $0.463 \pm 0.040$ & $0.489 \pm 0.028$ & $0.493 \pm 0.031$
% & $0.497 \pm 0.028$ \\
% \method-He & SEAN-T+SNS
% & $0.362 \pm 0.058$
% & $0.447 \pm 0.051$
% & $0.470 \pm 0.044$
% & $0.493 \pm 0.032$
% & $0.501 \pm 0.028$
% & $0.503 \pm 0.022$ \\
% \midrule
% \method-Ho & SEAN-T only
% & $0.274 \pm 0.045$
% & $0.332 \pm 0.036$
% & $0.351 \pm 0.034$
% & $0.365 \pm 0.030$
% & $0.407 \pm 0.024$
% & $0.419 \pm 0.010$ \\
% \method-Ho & SEAN-T+SNS
% & $0.344 \pm 0.050$
% & $0.431 \pm 0.045$
% & $0.453 \pm 0.035$
% & $0.468 \pm 0.030$
% & $0.471 \pm 0.024$
% & $0.475 \pm 0.020$ \\
% \bottomrule
% \end{tabular}
}
\end{table*}

\begin{table*}[t]
\centering
\small
\caption{Cross-dataset transfer to SEAN-T. Models are pretrained either
on SNS only or jointly on SEAN-T and SNS and evaluated on the three
SEAN-T perception tasks using progressively larger fractions of labeled
training data. Results report $F_1$-Score ($\mu \pm \sigma$).}
\label{tab:cross_dataset_sean}
\setlength{\tabcolsep}{4pt}
\resizebox{\textwidth}{!}{
\begin{tabular}{lllcccccc}
\toprule
\textbf{Model} & \textbf{Pretraining} & \textbf{Task} &
\textbf{1\%} & \textbf{5\%} & \textbf{10\%} &
\textbf{25\%} & \textbf{50\%} & \textbf{100\%} \\
\midrule

\method-He & SNS only & Competence
& $0.615 \pm 0.10$ & $0.674 \pm 0.08$ & $0.731 \pm 0.05$ & $0.772 \pm 0.03$ & $0.786 \pm 0.03$
& $0.801 \pm 0.01$ \\
\method-He & SNS only & Surprise
& $0.409 \pm 0.20$ & $0.446 \pm 0.18$ & $0.493 \pm 0.17$ & $0.595 \pm 0.11$ & $0.652 \pm 0.08$
& $0.710 \pm 0.05$ \\
\method-He & SNS only & Intention
& $0.571 \pm 0.09$ & $0.619 \pm 0.08$ & $0.668 \pm 0.04$ & $0.700 \pm 0.03$ & $0.717 \pm 0.03$
& $0.729 \pm 0.03$ \\

\method-He & SEAN-T+SNS & Competence
& $0.635 \pm 0.10$
& $0.673 \pm 0.07$
& $0.726 \pm 0.04$
& $0.762 \pm 0.03$
& $0.781 \pm 0.03$
& $0.794 \pm 0.02$ \\
\method-He & SEAN-T+SNS & Surprise
& $0.455 \pm 0.17$
& $0.481 \pm 0.17$
& $0.526 \pm 0.15$
& $0.603 \pm 0.12$
& $0.655 \pm 0.08$
& $0.709 \pm 0.04$ \\
\method-He & SEAN-T+SNS & Intention
& $0.576 \pm 0.08$
& $0.619 \pm 0.05$
& $0.671 \pm 0.04$
& $0.694 \pm 0.03$
& $0.705 \pm 0.03$
& $0.721 \pm 0.03$ \\

\midrule

\method-Ho & SNS only & Competence
& $0.440 \pm 0.27$
& $0.507 \pm 0.28$
& $0.600 \pm 0.24$
& $0.735 \pm 0.02$
& $0.744 \pm 0.03$
& $0.766 \pm 0.01$ \\
\method-Ho & SNS only & Surprise
& $0.206 \pm 0.28$
& $0.285 \pm 0.25$
& $0.309 \pm 0.26$
& $0.481 \pm 0.23$
& $0.570 \pm 0.14$
& $0.655 \pm 0.04$ \\
\method-Ho & SNS only & Intention
& $0.493 \pm 0.18$
& $0.515 \pm 0.19$
& $0.573 \pm 0.11$
& $0.647 \pm 0.02$
& $0.655 \pm 0.02$
& $0.676 \pm 0.02$ \\

\method-Ho & SEAN-T+SNS & Competence
& $0.612 \pm 0.14$
& $0.702 \pm 0.08$
& $0.738 \pm 0.04$
& $0.765 \pm 0.04$
& $0.787 \pm 0.03$
& $0.799 \pm 0.02$ \\
\method-Ho & SEAN-T+SNS & Surprise
& $0.440 \pm 0.23$
& $0.541 \pm 0.21$
& $0.548 \pm 0.20$
& $0.610 \pm 0.16$
& $0.713 \pm 0.06$
& $0.741 \pm 0.03$ \\
\method-Ho & SEAN-T+SNS & Intention
& $0.603 \pm 0.10$
& $0.671 \pm 0.07$
& $0.701 \pm 0.06$
& $0.715 \pm 0.04$
& $0.723 \pm 0.03$
& $0.734 \pm 0.02$ \\

\bottomrule
\end{tabular}
}
\end{table*}

At 100\% labeled data, \method-He achieves performance comparable to joint pretraining. SEAN-T-only pretraining achieves
$0.497$ macro $F_1$ on SNS compared with $0.503$ under joint
pretraining, while SNS-only pretraining achieves
$0.801/0.710/0.729$ on SEAN-T competence, surprise, and intention,
compared with $0.794/0.709/0.721$ under joint pretraining.
\method-Ho also transfers across datasets, although joint pretraining
provides larger gains.

\subsection{Pretraining Objective Ablations}
\label{sec:pretraining_objective_ablations}
\paragraph{Random Encoder.}
The random encoder baseline uses the same heterogeneous GNN-VAE architecture as the full model but removes self-supervised pretraining. This isolates whether downstream performance comes from the learned pretrained representation or simply from the architecture and linear probe. The random encoder performs noticeably worse than the full model on most of the F1-scores, indicating that the self-supervised graph reconstruction objective learns task-relevant latent structure.

\subsubsection{Self-Supervised vs. Supervised Heterogeneous Graph}
\label{sec:supervised_heterograph_ablation}
% new paragraph + table/figure

To separate the value of self-supervised pretraining from the heterogeneous graph architecture itself, we compare the frozen \method encoder with an otherwise similar heterogeneous model trained end-to-end using supervised SNS labels. As shown in Table~\ref{tbl:sns-sup-vs-selfsup}, supervised training improves some minority-class scores such as \emph{avoid}, but the self-supervised representation achieves stronger macro-$F_1$ across all label fractions. This suggests that the reconstruction-based pretraining objective provides a more stable representation for low-data downstream adaptation under the SNS class imbalance.

\begin{table*}[t]
  \centering
  \small
  \caption{SNS Action Classification: Self-Supervised vs. Supervised GNN-VAE. We compare STARS' frozen self-supervised encoder with a linear probe against an identical
  architecture trained end-to-end with supervised classification. Both models use the heterogeneous graph formulation. Results report F1-Score ($\mu \pm \sigma$) across
  label fractions.}   
  \label{tbl:sns-sup-vs-selfsup}
  \resizebox{\textwidth}{!}{
\begin{tabular}{llcccccc}
  \toprule
  \textbf{Task} & \textbf{Method} & \textbf{1\%} & \textbf{5\%} & \textbf{10\%} & \textbf{25\%} & \textbf{50\%} & \textbf{100\%} \\
  \midrule
  \multirow{2}{*}{Avoid} & Hetero Self-Sup & $0.098 \pm 0.09$ & $0.173 \pm 0.11$ & $0.190 \pm 0.10$ & $0.219 \pm 0.10$ & $0.230 \pm 0.09$ & $0.242 \pm 0.08$ \\
   & Hetero Sup & $0.178 \pm 0.07$ & $0.234 \pm 0.05$ & $0.258 \pm 0.04$ & $0.292 \pm 0.03$ & $0.297 \pm 0.02$ & $0.284 \pm 0.02$ \\
  \midrule
  \multirow{2}{*}{Follow} & Hetero Self-Sup & $0.155 \pm 0.15$ & $0.327 \pm 0.12$ & $0.374 \pm 0.08$ & $0.404 \pm 0.04$ & $0.412 \pm 0.03$ & $0.405 \pm 0.03$ \\
   & Hetero Sup & $0.112 \pm 0.11$ & $0.287 \pm 0.07$ & $0.305 \pm 0.06$ & $0.353 \pm 0.04$ & $0.373 \pm 0.04$ & $0.384 \pm 0.04$ \\
  \midrule
  \multirow{2}{*}{Not Consider} & Hetero Self-Sup & $0.834 \pm 0.06$ & $0.840 \pm 0.04$ & $0.846 \pm 0.03$ & $0.855 \pm 0.02$ & $0.861 \pm 0.02$ & $0.862 \pm 0.02$ \\
   & Hetero Sup & $0.218 \pm 0.25$ & $0.351 \pm 0.22$ & $0.429 \pm 0.23$ & $0.604 \pm 0.13$ & $0.662 \pm 0.08$ & $0.667 \pm 0.07$ \\
  \midrule
  \multirow{2}{*}{Macro $F_1$} & Hetero Self-Sup & $0.362 \pm 0.06$ & $0.447 \pm 0.05$ & $0.470 \pm 0.04$ & $0.493 \pm 0.03$ & $0.501 \pm 0.03$ & $0.503 \pm 0.02$ \\
   & Hetero Sup & $0.169 \pm 0.09$ & $0.291 \pm 0.10$ & $0.331 \pm 0.10$ & $0.417 \pm 0.07$ & $0.444 \pm 0.04$ & $0.445 \pm 0.04$ \\
  \bottomrule
\end{tabular}
  % \begin{tabular}{llcccccc}
  % \toprule
  % \textbf{Task} & \textbf{Method} & \textbf{1\%} & \textbf{5\%} & \textbf{10\%} & \textbf{25\%} & \textbf{50\%} & \textbf{100\%} \\
  % \midrule
  % \multirow{2}{*}{Avoid} & Hetero Self-Sup & $0.098 \pm 0.087$ & $0.173 \pm 0.107$ & $0.190 \pm 0.104$ & $0.219 \pm 0.097$ & $0.230 \pm 0.088$ & $0.242 \pm 0.078$ \\
  %  & Hetero Sup & $0.178 \pm 0.067$ & $0.234 \pm 0.047$ & $0.258 \pm 0.038$ & $0.292 \pm 0.032$ & $0.297 \pm 0.024$ & $0.284 \pm 0.019$ \\
  % \midrule
  % \multirow{2}{*}{Follow} & Hetero Self-Sup & $0.155 \pm 0.150$ & $0.327 \pm 0.115$ & $0.374 \pm 0.084$ & $0.404 \pm 0.036$ & $0.412 \pm 0.034$ & $0.405 \pm 0.030$ \\
  %  & Hetero Sup & $0.112 \pm 0.113$ & $0.287 \pm 0.065$ & $0.305 \pm 0.059$ & $0.353 \pm 0.044$ & $0.373 \pm 0.035$ & $0.384 \pm 0.045$ \\
  % \midrule
  % \multirow{2}{*}{Not Consider} & Hetero Self-Sup & $0.834 \pm 0.056$ & $0.840 \pm 0.038$ & $0.846 \pm 0.033$ & $0.855 \pm 0.024$ & $0.861 \pm 0.021$ & $0.862 \pm 0.019$
  % \\
  %  & Hetero Sup & $0.218 \pm 0.247$ & $0.351 \pm 0.222$ & $0.429 \pm 0.227$ & $0.604 \pm 0.134$ & $0.662 \pm 0.083$ & $0.667 \pm 0.072$ \\
  % \midrule
  % \multirow{2}{*}{Macro $F_1$} & Hetero Self-Sup & $0.362 \pm 0.058$ & $0.447 \pm 0.051$ & $0.470 \pm 0.044$ & $0.493 \pm 0.032$ & $0.501 \pm 0.028$ & $0.503 \pm 0.022$ \\
  %  & Hetero Sup & $0.169 \pm 0.093$ & $0.291 \pm 0.103$ & $0.331 \pm 0.103$ & $0.417 \pm 0.066$ & $0.444 \pm 0.044$ & $0.445 \pm 0.040$ \\
  % \bottomrule
  % \end{tabular}
  }
  \end{table*}

% end section

\begin{table*}[t]
\centering
\small
\caption{Random encoder ablation results.}
\label{tab:random_encoder_ablation}
\resizebox{\textwidth}{!}{
\begin{tabular}{lllcccccc}
\toprule
 & & \textbf{Architecture} & \textbf{1\%} & \textbf{5\%} & \textbf{10\%} & \textbf{25\%} & \textbf{50\%} & \textbf{100\%} \\
\midrule
Random encoder: & Competence & Heterogeneous & $0.530 \pm 0.22$ & $0.596 \pm 0.16$ & $0.635 \pm 0.12$ & $0.677 \pm 0.09$ & $0.691 \pm 0.08$ & $0.707 \pm 0.07$ \\
 & Surprise & Heterogeneous & $0.251 \pm 0.20$ & $0.296 \pm 0.21$ & $0.372 \pm 0.17$ & $0.424 \pm 0.13$ & $0.467 \pm 0.12$ & $0.480 \pm 0.14$ \\
 & Intention & Heterogeneous & $0.606 \pm 0.20$ & $0.648 \pm 0.14$ & $0.677 \pm 0.10$ & $0.700 \pm 0.08$ & $0.704 \pm 0.08$ & $0.713 \pm 0.09$ \\
 & $F_1$ Avoid & Heterogeneous & $0.070 \pm 0.11$ & $0.251 \pm 0.08$ & $0.301 \pm 0.03$ & $0.304 \pm 0.02$ & $0.305 \pm 0.02$ & $0.302 \pm 0.02$ \\
 & $F_1$ Follow & Heterogeneous & $0.070 \pm 0.12$ & $0.065 \pm 0.11$ & $0.101 \pm 0.12$ & $0.101 \pm 0.11$ & $0.082 \pm 0.10$ & $0.095 \pm 0.10$ \\
 & $F_1$ Not Consider & Heterogeneous & $0.865 \pm 0.03$ & $0.849 \pm 0.02$ & $0.833 \pm 0.02$ & $0.839 \pm 0.01$ & $0.839 \pm 0.01$ & $0.839 \pm 0.00$ \\
 & PA & Heterogeneous & $0.601 \pm 0.03$ & $0.590 \pm 0.02$ & $0.577 \pm 0.02$ & $0.583 \pm 0.01$ & $0.584 \pm 0.01$ & $0.583 \pm 0.00$ \\
\bottomrule
\end{tabular}
% \begin{tabular}{lllcccccc}
% \toprule
%  & & \textbf{Architecture} & \textbf{1\%} & \textbf{5\%} & \textbf{10\%} & \textbf{25\%} & \textbf{50\%} & \textbf{100\%} \\
% \midrule
% Random encoder: & Competence & Heterogeneous & $0.530 \pm 0.223$ & $0.596 \pm 0.159$ & $0.635 \pm 0.120$ & $0.677 \pm 0.090$ & $0.691 \pm 0.080$ & $0.707 \pm 0.074$ \\
%  & Surprise & Heterogeneous & $0.251 \pm 0.197$ & $0.296 \pm 0.206$ & $0.372 \pm 0.172$ & $0.424 \pm 0.133$ & $0.467 \pm 0.116$ & $0.480 \pm 0.142$ \\
%  & Intention & Heterogeneous & $0.606 \pm 0.200$ & $0.648 \pm 0.139$ & $0.677 \pm 0.100$ & $0.700 \pm 0.078$ & $0.704 \pm 0.080$ & $0.713 \pm 0.086$ \\
%  & $F_1$ Avoid & Heterogeneous & $0.070 \pm 0.109$ & $0.251 \pm 0.084$ & $0.301 \pm 0.026$ & $0.304 \pm 0.016$ & $0.305 \pm 0.018$ & $0.302 \pm 0.017$ \\
%  & $F_1$ Follow & Heterogeneous & $0.070 \pm 0.120$ & $0.065 \pm 0.110$ & $0.101 \pm 0.121$ & $0.101 \pm 0.113$ & $0.082 \pm 0.100$ & $0.095 \pm 0.102$ \\
%  & $F_1$ Not Consider & Heterogeneous & $0.865 \pm 0.028$ & $0.849 \pm 0.022$ & $0.833 \pm 0.021$ & $0.839 \pm 0.009$ & $0.839 \pm 0.006$ & $0.839 \pm 0.004$ \\
%  & PA & Heterogeneous & $0.601 \pm 0.026$ & $0.590 \pm 0.020$ & $0.577 \pm 0.020$ & $0.583 \pm 0.009$ & $0.584 \pm 0.006$ & $0.583 \pm 0.004$ \\
% \bottomrule
% \end{tabular}
}
\end{table*}

\subsection{Representation Component Ablations}
\label{sec:representation_component_ablations}

\paragraph{Robot-Latent Only.}
The robot-latent only ablation uses only the learned robot latent representation for downstream prediction, omitting human-node latents from the downstream readout.
This tests how much task-relevant information is summarized directly in the robot representation.
The full representation performs substantially better on SNS minority action classes and on most SEAN-T perception tasks, although the robot-latent only representation remains competitive on SEAN-T intention.

\begin{table*}[t]
\centering
\small
\caption{Robot-latent only (128D) ablation results.}
\label{tab:robot_z_only_ablation}
\resizebox{\textwidth}{!}{
\begin{tabular}{lllcccccc}
\toprule
 & & \textbf{Architecture} & \textbf{1\%} & \textbf{5\%} & \textbf{10\%} & \textbf{25\%} & \textbf{50\%} & \textbf{100\%} \\
\midrule
128D robot-z only & Competence & Heterogeneous & $0.592 \pm 0.13$ & $0.627 \pm 0.10$ & $0.663 \pm 0.08$ & $0.695 \pm 0.07$ & $0.707 \pm 0.07$ & $0.711 \pm 0.08$ \\
 & Surprise & Heterogeneous & $0.373 \pm 0.12$ & $0.400 \pm 0.12$ & $0.410 \pm 0.11$ & $0.432 \pm 0.12$ & $0.454 \pm 0.12$ & $0.479 \pm 0.14$ \\
 & Intention & Heterogeneous & $0.623 \pm 0.12$ & $0.662 \pm 0.08$ & $0.696 \pm 0.08$ & $0.716 \pm 0.07$ & $0.730 \pm 0.07$ & $0.730 \pm 0.07$ \\
 & $F_1$ Avoid & Heterogeneous & $0.017 \pm 0.05$ & $0.004 \pm 0.02$ & $0.001 \pm 0.01$ & $0.000 \pm 0.00$ & $0.000 \pm 0.00$ & $0.000 \pm 0.00$ \\
 & $F_1$ Follow & Heterogeneous & $0.004 \pm 0.02$ & $0.008 \pm 0.03$ & $0.007 \pm 0.02$ & $0.003 \pm 0.01$ & $0.000 \pm 0.00$ & $0.000 \pm 0.00$ \\
 & $F_1$ Not Consider & Heterogeneous & $0.875 \pm 0.02$ & $0.881 \pm 0.01$ & $0.883 \pm 0.00$ & $0.883 \pm 0.00$ & $0.883 \pm 0.00$ & $0.883 \pm 0.00$ \\
 & PA & Heterogeneous & $0.610 \pm 0.02$ & $0.616 \pm 0.01$ & $0.618 \pm 0.00$ & $0.618 \pm 0.00$ & $0.618 \pm 0.00$ & $0.618 \pm 0.00$ \\
\bottomrule
\end{tabular}
% \begin{tabular}{lllcccccc}
% \toprule
%  & & \textbf{Architecture} & \textbf{1\%} & \textbf{5\%} & \textbf{10\%} & \textbf{25\%} & \textbf{50\%} & \textbf{100\%} \\
% \midrule
% 128D robot-z only & Competence & Heterogeneous & $0.592 \pm 0.127$ & $0.627 \pm 0.102$ & $0.663 \pm 0.083$ & $0.695 \pm 0.072$ & $0.707 \pm 0.071$ & $0.711 \pm 0.081$ \\
%  & Surprise & Heterogeneous & $0.373 \pm 0.119$ & $0.400 \pm 0.124$ & $0.410 \pm 0.112$ & $0.432 \pm 0.124$ & $0.454 \pm 0.123$ & $0.479 \pm 0.143$ \\
%  & Intention & Heterogeneous & $0.623 \pm 0.116$ & $0.662 \pm 0.084$ & $0.696 \pm 0.080$ & $0.716 \pm 0.070$ & $0.730 \pm 0.071$ & $0.730 \pm 0.065$ \\
%  & $F_1$ Avoid & Heterogeneous & $0.017 \pm 0.046$ & $0.004 \pm 0.018$ & $0.001 \pm 0.008$ & $0.000 \pm 0.000$ & $0.000 \pm 0.000$ & $0.000 \pm 0.000$ \\
%  & $F_1$ Follow & Heterogeneous & $0.004 \pm 0.016$ & $0.008 \pm 0.034$ & $0.007 \pm 0.022$ & $0.003 \pm 0.015$ & $0.000 \pm 0.000$ & $0.000 \pm 0.000$ \\
%  & $F_1$ Not Consider & Heterogeneous & $0.875 \pm 0.022$ & $0.881 \pm 0.007$ & $0.883 \pm 0.001$ & $0.883 \pm 0.001$ & $0.883 \pm 0.000$ & $0.883 \pm 0.000$ \\
%  & PA & Heterogeneous & $0.610 \pm 0.021$ & $0.616 \pm 0.008$ & $0.618 \pm 0.001$ & $0.618 \pm 0.000$ & $0.618 \pm 0.000$ & $0.618 \pm 0.000$ \\
% \bottomrule
% \end{tabular}
}
\end{table*}

\paragraph{Bystander-Latent Only.}
The bystander-latent only ablation uses only the mean pooled bystander latent representation, omitting the robot latent from the downstream readout.
This tests how much predictive information can be recovered from surrounding pedestrian context alone.
The bystander representation retains task relevant signal, but the full representation performs better overall, indicating that combining robot and pedestrian representations  is more informative for downstream tasks.

\begin{table*}[t]
\centering
\small
\caption{128D byst only ablation results.}
\label{tab:byst_only_ablation}
\resizebox{\textwidth}{!}{
\begin{tabular}{lllcccccc}
\toprule
 & & \textbf{Architecture} & \textbf{1\%} & \textbf{5\%} & \textbf{10\%} & \textbf{25\%} & \textbf{50\%} & \textbf{100\%} \\
\midrule
128D byst only & Competence & Heterogeneous & $0.551 \pm 0.17$ & $0.533 \pm 0.15$ & $0.556 \pm 0.15$ & $0.621 \pm 0.10$ & $0.642 \pm 0.09$ & $0.650 \pm 0.07$ \\
 & Surprise & Heterogeneous & $0.323 \pm 0.15$ & $0.337 \pm 0.13$ & $0.302 \pm 0.13$ & $0.232 \pm 0.14$ & $0.161 \pm 0.15$ & $0.149 \pm 0.14$ \\
 & Intention & Heterogeneous & $0.574 \pm 0.17$ & $0.598 \pm 0.15$ & $0.634 \pm 0.13$ & $0.693 \pm 0.09$ & $0.713 \pm 0.09$ & $0.717 \pm 0.09$ \\
 & $F_1$ Avoid & Heterogeneous & $0.134 \pm 0.10$ & $0.154 \pm 0.08$ & $0.167 \pm 0.08$ & $0.149 \pm 0.09$ & $0.148 \pm 0.08$ & $0.153 \pm 0.09$ \\
 & $F_1$ Follow & Heterogeneous & $0.223 \pm 0.12$ & $0.317 \pm 0.06$ & $0.341 \pm 0.04$ & $0.344 \pm 0.03$ & $0.344 \pm 0.03$ & $0.343 \pm 0.03$ \\
 & $F_1$ Not Consider & Heterogeneous & $0.661 \pm 0.08$ & $0.756 \pm 0.06$ & $0.786 \pm 0.04$ & $0.807 \pm 0.03$ & $0.807 \pm 0.03$ & $0.811 \pm 0.02$ \\
 & PA & Heterogeneous & $0.419 \pm 0.06$ & $0.498 \pm 0.05$ & $0.524 \pm 0.03$ & $0.539 \pm 0.03$ & $0.539 \pm 0.02$ & $0.542 \pm 0.02$ \\
\bottomrule
\end{tabular}
% \begin{tabular}{lllcccccc}
% \toprule
%  & & \textbf{Architecture} & \textbf{1\%} & \textbf{5\%} & \textbf{10\%} & \textbf{25\%} & \textbf{50\%} & \textbf{100\%} \\
% \midrule
% 128D byst only & Competence & Heterogeneous & $0.551 \pm 0.166$ & $0.533 \pm 0.155$ & $0.556 \pm 0.150$ & $0.621 \pm 0.102$ & $0.642 \pm 0.087$ & $0.650 \pm 0.074$ \\
%  & Surprise & Heterogeneous & $0.323 \pm 0.147$ & $0.337 \pm 0.127$ & $0.302 \pm 0.126$ & $0.232 \pm 0.139$ & $0.161 \pm 0.146$ & $0.149 \pm 0.140$ \\
%  & Intention & Heterogeneous & $0.574 \pm 0.169$ & $0.598 \pm 0.148$ & $0.634 \pm 0.127$ & $0.693 \pm 0.093$ & $0.713 \pm 0.090$ & $0.717 \pm 0.087$ \\
%  & $F_1$ Avoid & Heterogeneous & $0.134 \pm 0.102$ & $0.154 \pm 0.084$ & $0.167 \pm 0.080$ & $0.149 \pm 0.086$ & $0.148 \pm 0.083$ & $0.153 \pm 0.087$ \\
%  & $F_1$ Follow & Heterogeneous & $0.223 \pm 0.119$ & $0.317 \pm 0.058$ & $0.341 \pm 0.035$ & $0.344 \pm 0.030$ & $0.344 \pm 0.031$ & $0.343 \pm 0.029$ \\
%  & $F_1$ Not Consider & Heterogeneous & $0.661 \pm 0.078$ & $0.756 \pm 0.062$ & $0.786 \pm 0.039$ & $0.807 \pm 0.030$ & $0.807 \pm 0.028$ & $0.811 \pm 0.022$ \\
%  & PA & Heterogeneous & $0.419 \pm 0.055$ & $0.498 \pm 0.047$ & $0.524 \pm 0.029$ & $0.539 \pm 0.025$ & $0.539 \pm 0.023$ & $0.542 \pm 0.019$ \\
% \bottomrule
% \end{tabular}
}
\end{table*}

\paragraph{Raw Robot Node Only.}
The raw robot-node baseline bypasses graph pretraining entirely and trains directly on the 592-dimensional SEAN robot feature vector. This tests whether the SEAN labels can be predicted from robot-centric observations alone without relational message passing. The raw robot-node baseline contains substantial predictive signal, particularly for competence and surprise.
At 100\% labeled data, it achieves $0.746/0.700/0.688$ on competence, surprise, and intention, compared with $0.794/0.709/0.721$ for the full heterogeneous \method representation.
Compared with the robot-latent-only ablation, the raw features perform better on competence and surprise but worse on intention, indicating that the different representations retain complementary task-relevant information.

\begin{table*}[t]
\centering
\small
\caption{592D robot node only (SEAN-ablation compared with combined model).}
\label{tab:robot_node_only_ablation}
\resizebox{\textwidth}{!}{
\begin{tabular}{llcccccc}
\toprule
 & \textbf{Architecture} & \textbf{1\%} & \textbf{5\%} & \textbf{10\%} & \textbf{25\%} & \textbf{50\%} & \textbf{100\%} \\
\midrule
Competence & Heterogeneous & $0.541 \pm 0.11$ & $0.641 \pm 0.07$ & $0.673 \pm 0.07$ & $0.714 \pm 0.05$ & $0.724 \pm 0.04$ & $0.746 \pm 0.03$ \\
Surprise & Heterogeneous & $0.528 \pm 0.10$ & $0.591 \pm 0.09$ & $0.610 \pm 0.08$ & $0.652 \pm 0.05$ & $0.682 \pm 0.04$ & $0.700 \pm 0.04$ \\
Intention & Heterogeneous & $0.498 \pm 0.09$ & $0.580 \pm 0.06$ & $0.590 \pm 0.05$ & $0.632 \pm 0.05$ & $0.649 \pm 0.05$ & $0.688 \pm 0.05$ \\
\bottomrule
\end{tabular}
% \begin{tabular}{llcccccc}
% \toprule
%  & \textbf{Architecture} & \textbf{1\%} & \textbf{5\%} & \textbf{10\%} & \textbf{25\%} & \textbf{50\%} & \textbf{100\%} \\
% \midrule
% Competence & Heterogeneous & $0.541 \pm 0.112$ & $0.641 \pm 0.075$ & $0.673 \pm 0.065$ & $0.714 \pm 0.047$ & $0.724 \pm 0.042$ & $0.746 \pm 0.032$ \\
% Surprise & Heterogeneous & $0.528 \pm 0.098$ & $0.591 \pm 0.087$ & $0.610 \pm 0.079$ & $0.652 \pm 0.052$ & $0.682 \pm 0.044$ & $0.700 \pm 0.040$ \\
% Intention & Heterogeneous & $0.498 \pm 0.087$ & $0.580 \pm 0.063$ & $0.590 \pm 0.054$ & $0.632 \pm 0.049$ & $0.649 \pm 0.050$ & $0.688 \pm 0.050$ \\
% \bottomrule
% \end{tabular}
}
\end{table*}

\subsection{Baseline and Lower-Bound Analyses}
\label{sec:baseline_lower_bound_analyses}

\paragraph{Weighted Random Baseline.}
The weighted random baseline samples predictions according to the empirical training-label distribution, providing a label-aware lower bound under class imbalance. Its strong performance on majority labels such as \emph{not consider} highlights how misleading accuracy or single-class F1 can be on imbalanced SNS labels. However, overall, our full model substantially outperforms this baseline in nearly every F1 metric, showing that it learns beyond label-frequency priors.

\begin{table*}[t]
\centering
\small
\caption{WRB ablation results.}
\label{tab:wrb_ablation}
\resizebox{\textwidth}{!}{
\begin{tabular}{lllcccccc}
\toprule
 & & \textbf{Architecture} & \textbf{1\%} & \textbf{5\%} & \textbf{10\%} & \textbf{25\%} & \textbf{50\%} & \textbf{100\%} \\
\midrule
WRB: & Competence & Heterogeneous & $0.524 \pm 0.14$ & $0.528 \pm 0.12$ & $0.517 \pm 0.10$ & $0.517 \pm 0.10$ & $0.530 \pm 0.11$ & $0.513 \pm 0.10$ \\
 & Surprise & Heterogeneous & $0.345 \pm 0.13$ & $0.366 \pm 0.12$ & $0.384 \pm 0.10$ & $0.360 \pm 0.08$ & $0.368 \pm 0.08$ & $0.360 \pm 0.09$ \\
 & Intention & Heterogeneous & $0.563 \pm 0.11$ & $0.574 \pm 0.09$ & $0.575 \pm 0.09$ & $0.578 \pm 0.08$ & $0.575 \pm 0.08$ & $0.563 \pm 0.08$ \\
 & $F_1$ Avoid & Heterogeneous & $0.090 \pm 0.07$ & $0.104 \pm 0.05$ & $0.104 \pm 0.04$ & $0.108 \pm 0.04$ & $0.105 \pm 0.05$ & $0.103 \pm 0.05$ \\
 & $F_1$ Follow & Heterogeneous & $0.046 \pm 0.06$ & $0.062 \pm 0.05$ & $0.069 \pm 0.05$ & $0.083 \pm 0.04$ & $0.070 \pm 0.05$ & $0.083 \pm 0.05$ \\
 & $F_1$ Not Consider & Heterogeneous & $0.812 \pm 0.05$ & $0.806 \pm 0.03$ & $0.800 \pm 0.02$ & $0.811 \pm 0.02$ & $0.813 \pm 0.02$ & $0.809 \pm 0.01$ \\
 & PA & Heterogeneous & $0.542 \pm 0.05$ & $0.532 \pm 0.03$ & $0.527 \pm 0.02$ & $0.536 \pm 0.02$ & $0.539 \pm 0.01$ & $0.536 \pm 0.01$ \\
\bottomrule
\end{tabular}
% \begin{tabular}{lllcccccc}
% \toprule
%  & & \textbf{Architecture} & \textbf{1\%} & \textbf{5\%} & \textbf{10\%} & \textbf{25\%} & \textbf{50\%} & \textbf{100\%} \\
% \midrule
% WRB: & Competence & Heterogeneous & $0.524 \pm 0.145$ & $0.528 \pm 0.119$ & $0.517 \pm 0.096$ & $0.517 \pm 0.100$ & $0.530 \pm 0.107$ & $0.513 \pm 0.102$ \\
%  & Surprise & Heterogeneous & $0.345 \pm 0.130$ & $0.366 \pm 0.119$ & $0.384 \pm 0.099$ & $0.360 \pm 0.082$ & $0.368 \pm 0.084$ & $0.360 \pm 0.089$ \\
%  & Intention & Heterogeneous & $0.563 \pm 0.114$ & $0.574 \pm 0.093$ & $0.575 \pm 0.089$ & $0.578 \pm 0.083$ & $0.575 \pm 0.082$ & $0.563 \pm 0.081$ \\
%  & $F_1$ Avoid & Heterogeneous & $0.090 \pm 0.066$ & $0.104 \pm 0.048$ & $0.104 \pm 0.045$ & $0.108 \pm 0.041$ & $0.105 \pm 0.046$ & $0.103 \pm 0.048$ \\
%  & $F_1$ Follow & Heterogeneous & $0.046 \pm 0.056$ & $0.062 \pm 0.047$ & $0.069 \pm 0.049$ & $0.083 \pm 0.043$ & $0.070 \pm 0.050$ & $0.083 \pm 0.052$ \\
%  & $F_1$ Not Consider & Heterogeneous & $0.812 \pm 0.049$ & $0.806 \pm 0.032$ & $0.800 \pm 0.023$ & $0.811 \pm 0.017$ & $0.813 \pm 0.016$ & $0.809 \pm 0.014$ \\
%  & PA & Heterogeneous & $0.542 \pm 0.050$ & $0.532 \pm 0.033$ & $0.527 \pm 0.023$ & $0.536 \pm 0.017$ & $0.539 \pm 0.014$ & $0.536 \pm 0.013$ \\
% \bottomrule
% \end{tabular}
}
\end{table*}

\subsection{SNS \texorpdfstring{$F_1$}--Score Analysis}
\label{sec:sns_f1_score_analysis}

We evaluate the SocialNav-SUB vision-language-model (VLM) baselines on the same held-out SNS test split used for our downstream action classification task. Rather than independently rerunning the prompted VLM baselines, we obtained their per-example prediction outputs from the SocialNav-SUB authors and computed metrics on the corresponding examples in our evaluation split. For each scenario, the model predicts a discrete relational action label for the target pedestrian. We compute per-class $F_1$-scores and Probability of Agreement (\textit{PA}) following the SocialNav-SUB evaluation protocol~\cite{munje2025socialnavsub}. Because these models are evaluated zero-shot through prompting and are not fine-tuned for this task, we report results only on the full test set rather than across labeled-data fractions. Table~\ref{sup:tab:sns_baseline} summarizes the resulting per-class $F_1$ and \textit{PA} scores.

\begin{table}[ht!p]
\centering
\caption{SocialNav-SUB VLM baseline results on the held-out SNS action classification task~\cite{munje2025socialnavsub}. Models are evaluated on the full downstream test set. Metrics report per-class $F_1$-scores alongside Probability of Agreement (\textit{PA}).}

\label{sup:tab:sns_baseline}
\setlength{\tabcolsep}{8pt} % Provides comfortable padding before scaling
\begin{tabular}{r *{6}{c}} % Keeps identical column layout structure to your main table
\toprule
\textbf{Model} &  $F_1$ \textbf{Avoid} & $F_1$ \textbf{Follow} & $F_1$ \textbf{Overtake} & $F_1$ \textbf{Yield} & $F_1$ \textbf{NC} & \textbf{PA} \\
\midrule
Gemini  & $0.223$ & $0.000$ & $0.000$ & $0.222$ & $0.745$ & $0.493$ \\
4o      & $0.241$ & $0.176$ & $0.000$ & $0.000$ & $0.521$ & $0.344$ \\
o4-mini & $0.222$ & $0.363$ & $0.000$ & $0.111$ & $0.852$ & $0.582$ \\
\bottomrule
\end{tabular}
\end{table}

\FloatBarrier
\clearpage

\end{document}